\documentclass{article}

\usepackage{microtype}
\usepackage{graphicx}
\usepackage{booktabs} % for professional tables

\usepackage{xr-hyper}
\usepackage{amssymb,dsfont,amsthm,amsmath}
\usepackage{mathtools}
\usepackage{xspace}
\usepackage{multirow}
\usepackage{listings}
\usepackage{bm}
\usepackage{subfig}
\usepackage{array}
\usepackage{eqparbox}
\usepackage{etoolbox}  % patch def of algorithmic environment

\usepackage{tikz}
\usetikzlibrary{shapes.geometric}
\usetikzlibrary{positioning,shapes,shadows,arrows,calc,automata}
\usetikzlibrary{intersections}
\pgfdeclarelayer{background}
\pgfdeclarelayer{foreground}
\pgfsetlayers{background,main,foreground}
\tikzstyle{activity}=[rectangle, draw=black, rounded corners, text centered, text width=6em, fill=white, drop shadow]
\tikzstyle{data}=[rectangle, draw=black, text centered, fill=black!10, text width=6em, drop shadow]
\tikzstyle{box}=[rectangle,draw=black,thick, minimum size=.5cm]
\tikzstyle{myarrow}=[->, thick]
\definecolor{my_magenta}{rgb}{0.796875,0.46875,0.734375}
\definecolor{my_green}{rgb}{0.0078125,0.6171875,0.44921875}
\definecolor{my_gold}{rgb}{0.8671875,0.55859375,0.01953125}
\definecolor{my_gray}{rgb}{0.5,0.5,0.5}
\definecolor{my_blue}{rgb}{0.4,0.4,1.}
\definecolor{my_ddpg_orange}{rgb}{0.7882352,0.560784,0.364705}
\definecolor{my_tddpg_blue}{rgb}{0.337254,0.70588,0.913725}
\definecolor{my_figar_gray}{rgb}{0.5803921,0.5803921,0.5803921}
\usepackage{todonotes}
\presetkeys{todonotes}{inline,backgroundcolor=red!40,caption={}}{}
\usepackage{wrapfig}

\usepackage{hyperref}
\usepackage{url}

\mathchardef\mhyphen="2D

\let\oldcite=\cite
\renewcommand\cite[1]{\ifthenelse{\equal{#1}{NEEDED}}{[{\color{blue}\texttt{citation~needed}}]}{\oldcite{#1}}}

\newcommand{\qlearning}{$\mathcal{Q}$-learning\xspace}
\newcommand{\Q}{$\mathcal{Q}$\xspace}
\newcommand{\q}{\mathcal{Q}}

\newcommand{\temporal}{{\textsc{tempoRL}}\xspace}  % TEMPOral Reinforcement Learning

\newcommand{\states}{\mathcal{S}}
\newcommand{\actions}{\mathcal{A}}
\newcommand{\transitions}{\mathcal{P}}  % I think in RL its common to have P as symbol for transition probabilities
\newcommand{\rewards}{\mathcal{R}}
\newcommand{\mdp}{\mathcal{M}}
\newcommand{\skips}{\mathcal{J}}

\newcommand{\skipaction}{j}

\DeclareMathOperator*{\egargmax}{-greedy\,arg\,max}

\theoremstyle{plain}

\theoremstyle{definition}

\theoremstyle{remark}

\newcommand{\denselist}{\itemsep -2pt\partopsep 0pt}

\usepackage{hyperref}

\usepackage[accepted]{icml2021}

\renewcommand\algorithmiccomment[1]{%
  \hfill\#\ \eqparbox{COMMENT}{#1}%
}
\newcommand\LONGCOMMENT[1]{%
  \hfill\#\ \begin{minipage}[t]{\eqboxwidth{COMMENT}}#1\strut\end{minipage}%
}

\icmltitlerunning{TempoRL: Learning When to Act}

\begin{document}

\twocolumn[
\icmltitle{TempoRL: Learning When to Act}



\icmlsetsymbol{equal}{*}

\begin{icmlauthorlist}
\icmlauthor{André Biedenkapp}{fr}
\icmlauthor{Raghu Rajan}{fr}
\icmlauthor{Frank Hutter}{fr,bo}
\icmlauthor{Marius Lindauer}{hn}
\end{icmlauthorlist}

\icmlaffiliation{fr}{Department  of  Computer  Science,  University  of  Freiburg, Germany}
\icmlaffiliation{bo}{BCAI, Renningen, Germany}
\icmlaffiliation{hn}{Information Processing Institute (tnt),  Leibniz University Hannover, Germany}

\icmlcorrespondingauthor{André Biedenkapp}{biedenka@cs.uni-freiburg.de}

\icmlkeywords{Temporal Abstraction, Temporal Decision Making, Skip-MDP, Reinforcement Learning}

\vskip 0.3in
]



\printAffiliationsAndNotice{}  % leave blank if no need to mention equal contribution

\begin{abstract}
Reinforcement learning is a powerful approach to learn behaviour through interactions with an environment.
However, behaviours are usually learned in a purely reactive fashion, where an appropriate action is selected based on an observation.
In this form, it is challenging to learn \emph{when} it is necessary to execute new decisions.
This makes learning inefficient, especially in environments that need various degrees of fine and coarse control.
To address this, we propose a proactive setting in which the agent not only selects an action in a state but also for how long to commit to that action.
Our TempoRL approach introduces skip connections between states and learns a skip-policy for repeating the same action along these skips.
We demonstrate the effectiveness of TempoRL on a variety of traditional and deep RL environments, showing that our approach is capable of learning successful policies up to an order of magnitude faster than vanilla \qlearning.
\end{abstract}

\section{Introduction}
Although reinforcement learning (RL) has celebrated many successes in the recent years \citep[see e.g.,][]{Mnih-nature15,lillicrap-iclr16,baker-iclr20}, in its classical form it is limited to learning policies in a mostly reactive fashion, i.e., observe a state and react to that state with an action.
Guided by the reward signal, policies that are learned in such a way can decide which action is expected to yield maximal long-term rewards.
However, these policies generally do not learn \emph{when a new decision has to be made}.
A more proactive way of learning, in which agents proactively commit to playing an action for multiple steps could further improve RL by (i) potentially providing better exploration compared to common one-step exploration; (ii) faster learning as proactive policies provide a form of temporal abstraction by requiring fewer decisions; (iii) explainability as learned agents can indicate when they expect new decisions are required.

Temporal abstractions are a common way to simplify learning of policies with potentially long action sequences. 
Typically, the temporal abstraction is learned on the highest level of a hierarchy and the required behaviour on a lower level~\citep[see e.g.][]{sutton-ai99,eysenbach-neurips19}.
For example, on the highest level a \textit{goal policy} learns which states are necessarily visited and on the lower level the \textit{behaviour} to reach goals is learned. Spacing goals far apart still requires to learn complex behaviour policies whereas a narrow goal spacing requires to learn complex goal policies. Another form of temporal abstraction is to use actions that work at different time-scales~\cite{precup-ecml98}.
Take for example an agent that is tasked with moving an object.
On the highest level the agent would follow a policy with abstract actions, such as \textit{pick-up object}, \textit{move object}, \textit{put-down object}, whereas on the lower level actions could directly control actuators to perform the abstract actions.

Such hierarchical approaches are still reactive, but instead of reacting to an observation on only one level, reactions are learned on multiple levels.
Though these approaches might allow us to learn \textit{which} states are necessarily traversed in the environment, they do not enable us to learn \textit{when} a new decision has to be made on the behaviour level.

In this work, we propose an alternative approach: a proactive view on learning policies that allows us to jointly learn a behaviour and how long to carry out that behaviour. To this end, we  re-examine the relationship between agent and environment, and the dependency on time. This allows us to introduce \textit{skip connections} for an environment. These skip connections do not change the optimal policy or state-action-values but allow us to propagate information much faster. We demonstrate the effectiveness of our method, which we dub \temporal with tabular and deep function approximation on a variety of environments with discrete and continuous action spaces.
Our contributions are:

\begin{enumerate}
\denselist
    \item We propose a proactive alternative to classical RL.
    \item We introduce skip-connections for MDPs by playing an action for several consecutive states, which leads to faster propagation of information about future rewards.
    \item We propose a mechanism based on a hierarchy for learning when to make new decisions through the use of skip-connections.
    \item On classical and deep RL benchmarks, we show that \temporal outperforms plain DQN, DAR and FiGAR both in terms of learning speed and sometimes even by converging to better policies.
\end{enumerate}

%--------------------------------------------------------------------------------------------
%--------------------------------------------------------------------------------------------
%--------------------------------------------------------------------------------------------
\section{Related Work}
A common framework for temporal abstraction in RL is the options framework \cite{precup-ecml98,sutton-ai99,stolle-sara02,bacon-aaai17,harutyunyan-aaai18,mankowitz-aaai18,khetrapal-aaai19}.
 Options are triples $\langle\mathcal{I},\pi,\beta\rangle$ where $\mathcal{I}$ is the set of admissible states that defines in which states the option can be played; $\pi$ is the policy the option follows when it is played; and $\beta$ is a random variable that determines when an option is terminated.
In contrast to our work, options require a lot of prior knowledge about the environment to determine the set of admissible states, as well as the option policies themselves.
However, \citet{chaganty-aamas12} proposed to learn options based on observed connectedness of states.
Similarly, SoRB~\cite{eysenbach-neurips19} uses data from the replay buffer to build a connectedness graph, which allows to query sub-goals on long trajectories.
Further work on discovering options paid attention to the termination criterion, learning persistent options~\cite{harb-aaai18-option-deliberation} and meaningful termination criteria~\cite{vezhnevets-neurips16,harutyunyan-aistats19-termination-critic}.

Similarly, in AI planning macro actions provide temporal abstractions.
However, macro actions are not always applicable as some actions can be locked.
\citet{chrpa-aaai19} propose to learn when macro actions become available again, allowing them to identify non-trivial activities.
For various problem domains of AI planning, varieties of useful macro actions are known and selecting which macro actions to consider is not trivial.
\citet{vallati-jetai19} propose a macro action selection mechanism that selects which macro actions should be considered for new problems.
Further, \citet{nasiriany-neurips19} show that goal-conditioned policies learned with RL can be incorporated into planning.
With complex state observations goal states are difficult to define.

An important element in DQN's success in tackling various Atari games~\cite{Mnih-nature15} is due to the use of \textit{frame skipping}~\cite{bellamare-jair13}.
Thereby the agent
skips over a few states, always playing the same action, before making a new decision.
Without the use of frame skipping, the change between successive observations is small and would have required more observations to learn the same policy.
Tuning the \textit{skip-size} can additionally improve performance \cite{braylan-l4gcvg15,khan-jr19}.
A similar line of research focuses on learning persistent policies which always act after a static, fixed time-step for one-dimensional~\cite{metelli-icml20-control-frequency-adaptation} and multi-dimensional actions~\cite{lee-neurips20-control-multi-frequency}.
However, static skip-sizes might not be ideal.
\citet{dabney-arxiv20} demonstrated that temporally extended $\epsilon$-greedy schedules improve exploration and thereby performance in sparse-reward environments while performing close to vanilla $\epsilon$-greedy exploration on dense-reward environments.

Different techniques have been proposed to handle continuous time environments \cite{doya-nc00,tiganj-sssaaai17}.
Recently, \citet{huang-allerton19} proposed to use \textit{Markov Jump Processes} (MJPs).
MJPs are designed to study optimal control in MDPs where observations have an associated cost. 
The goal then is to balance the costs of observations and actions to act in an optimal manner with respect to total cost.
Their analysis demonstrated that frequent observations are necessary in regions where an optimal action might change rapidly, while in areas of infrequent change, fewer observations are sufficient.
In contrast to ours, this formalism strictly prohibits observations of the skipped transitions to save observation costs and thus losing a lot of information, which otherwise could be used to learn how to act while simultaneously learning when new decisions are required.

\citet{schoknecht-icann02,schocknecht-nca03} demonstrated that learning with multi-step actions can significantly speed up RL.
Relatedly, \citet{lakshminarayanan-aaai17} proposed \emph{DAR}, a $\q$-network with multiple output heads per action to handle different repetition lengths, drastically increasing the action space but improving learning.
In contrast to that, 
\citet{sharma-iclr17} proposed \emph{FiGAR}, a framework that jointly learns an action policy and a second repetition policy that decides how often to repeat an action.
Crucially, its repetition policy is not conditioned on the chosen action resulting in independent repetition and behaviour actions.
The polices are learned together through a joint loss.
Thus, counter to our work, the repetition policy only learns which repetition length works well on average for all actions. 
Further, FiGAR requires modification to the training method of a base agent to accommodate the repetition policy.
When evaluating our method in the context of DQN, we compare against DAR and in the context of DDPG against FiGAR as they were originally developed and evaluated on these agent types.
The appendix, code and experiment results are available at \href{https://www.github.com/automl/TempoRL}{\nolinkurl{github.com/automl/TempoRL}}.
%--------------------------------------------------------------------------------------------
%--------------------------------------------------------------------------------------------
%--------------------------------------------------------------------------------------------
\section{TempoRL}
We begin this section by introducing skip connections into MDPs, propagating information about expected future rewards faster. We then introduce a novel learning mechanism that makes use of a hierarchy to learn a policy that is capable of not only learning which action to take, but also \textit{when} a new action has to be chosen.

\subsection{Temporal Abstraction through Skip MDPs}
It is possible to make use of contextual information in MDPs \citep[][]{hallak-corr15,modi-alt18,biedenkapp-ecai20}.
To this end, we contextualize an existing MDP $\mdp$ to allow for skip connections as $\mdp_\mathcal{C}\coloneqq\left\{\mdp_c\right\}_{c\in\mathcal{C}}$ with $\mdp_{c}\coloneqq\langle\states,\actions,\transitions_{c},\rewards_{c}\rangle$.
Akin to options, a skip-connection $c$ is a triple $\langle s, a, \skipaction\rangle$, where $s$ is the starting state for a skip transition (and not a set of states as in the options framework); $a$ is the action that is executed when skipping through the MDP; and $\skipaction$ is the skip-length, where $a\in\actions$, $s\in\states$ and $\skipaction\in\skips=\left\{1,\ldots,J\right\}$. 
This context to the MDP induces different MDPs with shared state and action spaces ($\states$, $\actions$), but different transitions $\transitions_c$ and reward functions $\rewards_c$ to account for the introduced skips.

In practice however, the transition and reward functions are unknown and do not allow to easily insert skips.
Nevertheless, as we make use of action repetition, we can simulate a skip connection.
A skip connects two states $s$ and $s'$ iff state $s'$ is reachable from state $s$ by repeating action $a$ $\skipaction$-times. This gives us the following skip transition function: \begin{equation}
    \transitions_c(s, a, s') = \left\{\begin{array}{ll}
         \prod_{k=0}^{\skipaction-1} \transitions_{s_ks_{k+1}}^a & \text{if reachable}\\
         0&  \text{otherwise}\\
    \end{array}\right.
\end{equation}
with $s_k$ and $s_{k+1}$ the states traversed by playing action $a$ for the $k$th time, and with $s_0=s$ and $s_{\skipaction}=s'$.
This change in the transition function is reflected accordingly in the reward: 
\begin{equation}
    \rewards_c(s, a, s') = \left\{
    \begin{array}{ll}
         \sum_{k=0}^{\skipaction-1} \gamma^k\rewards_{s_ks_{k+1}}^a&  \text{if reachable}\\
         0& \text{otherwise}.
    \end{array}
    \right.
\end{equation}
Thus, for skips of length $1$ we recover the original transition function $\transitions_{\langle s,a,1\rangle}(s,a,s') = \transitions_{ss'}^a$ as well as the original reward function $\rewards_{\langle s,a,1\rangle}(s,a,s') = \rewards_{ss'}^a$.
The goal with skip-MDPs is to find an optimal skip policy $\pi_J\colon\states\times\actions\mapsto\mathcal{J}$, i.e., a policy that takes a state and a behaviour action as input and maps to a skip value that maximally reduces the total required number of decisions  to reach the optimal reward.
Thus, similar to skip-connections in neural networks, skip MDPs allow us to propagate information about future rewards much more quickly and enables us to determine \emph{when} it becomes beneficial to switch actions.

\begin{figure}[tbp]
    \vskip -0.15in
    \centering
    \scalebox{.85}{
    \begin{tikzpicture}[node distance=1.25cm]
    	\node[state,inner sep=2pt,minimum size=0pt] (a) {$s_0$};
    	\node[state,right of=a,inner sep=2pt,minimum size=0pt] (b) {$s_{1}$};
    	\node[state,right of=b,inner sep=2pt,minimum size=0pt] (c) {$s_{2}$};
    	\node[state,right of=c,inner sep=2pt,minimum size=0pt] (d) {$s_{3}$};
    	
    	\node[above of=b] (aboveb) {};
    	\node[above of=c] (abovec) {};
    	\node[above of=d] (aboved) {};
    	
    	\path[every node/.style={font=\sffamily\scriptsize}]
    	(d) edge[red!50!black,<-,bend right=0] node[black,yshift=0.15cm] {} (c)
    	(c) edge[red!50!black,<-,bend right=0] node[black,yshift=0.15cm] {} (b)
    	(b) edge[red!50!black,<-,bend right=0] node[black,yshift=0.15cm] {} (a);

    	\path[every node/.style={font=\sffamily\scriptsize}]
    	(d) edge[red!0!black,<-,densely dotted,bend right=45] node[black,yshift=0.15cm] {} (a)
    	(d) edge[red!0!black,<-,dashed,bend right=30] node[black,yshift=0.15cm] {} (b)
    	(c) edge[red!0!black,<-,dashed,bend right=30] node[black,yshift=0.15cm] {} (a);
    \end{tikzpicture}
    }
    \caption{Example transitions with skip of length three (drawn with $\cdots$). At the same time we can also observe shorter skips of length two (-~-~-) and normal steps, i.e. skips of length one ($-$).}
    \label{fig:example_skip}
    \vskip -0.15in
\end{figure}
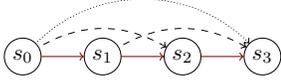
\subsection{Learning When to Make Decisions}
In order to learn using skip connections we need a new mechanism that selects which skip connection to use.
In order to facilitate this, we propose using a hierarchy in which a \textit{behaviour} policy determines the action $a$ to be played given the current state $s$, and a \textit{skip} policy determines how long to commit to this behaviour.

To learn the behaviour, we can make use of classical \qlearning, where the $\q$-function gives a mapping of expected future rewards when playing action $a$ in state $s_t$ at time $t$ and continuing to follow the behaviour policy $\pi$ thereafter.
\begin{equation}\label{eq:q}
    \q^\pi(s, a)\coloneqq\mathbb{E}\left[r_{t} + \gamma \q^\pi(s_{t+1}, a_{t+1})|s=s_t, a\right]
\end{equation}
To learn to skip, we first have to define a \textit{skip-action space} that determines all possible lengths of skip-connections, e.g., $j\in\left\{1,2,\ldots,J\right\}$. To learn the value of a skip we can make use of n-step \qlearning with the condition that, at each step of the $j$ steps, the action stays the same.
\begin{multline}\label{eq:skip}
    \q^{\pi_J}(s, j | a)\coloneqq\\
    \mathbb{E}\left[\sum_{k=0}^{j-1}\gamma^{k} r_{t+k}
    + \gamma^{j} \q^\pi(s_{t+\skipaction}, a_{t + \skipaction}) | s=s_t, a, j\right]
\end{multline}
We call this a flat hierarchy since the behaviour and the skip policy have to always make decisions at the same time-step; however, the behaviour policy has to be queried before the skip policy.
Once we have determined both the action $a$ and the skip-length $j$ we want to perform, we execute this action for $j$ steps.
We can then use standard temporal difference updates to update the behaviour and skip $\q$-functions with all one-step observations and the overarching skip-observation.
Note that the skip $\q$-function can also be conditioned on continuous actions if the behaviour policy can handle continuous action-spaces.

One interesting observation regarding this learning scheme is that, when playing skip action $\skipaction$, we are able to also observe all smaller skip transitions for all intermediate steps.
Figure~\ref{fig:example_skip} gives a visual representation.
Specifically, we can directly see that, when executing a skip of length $j$, we can observe and learn from $\frac{j\cdot(j + 1)}{2}$ skip-transitions in total.
As we observe all intermediate steps, we can use this trajectory of transitions to build a local connectedness graph (similar to Figure~\ref{fig:example_skip}) from which we can look up all skip-connections.
This allows us to efficiently learn the values of multiple skips, i.e. the action value at different time-resolutions.
For pseudo-code and more details we refer to Appendix~\ref{appendix:sec:implement}.

\subsection{Learning When to Make Decisions in Deep RL}\label{sec:temporl}%
\def\layersep{2.5cm}%
\begin{figure}[tbp]
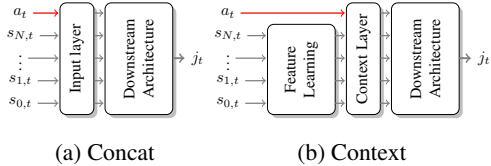
%
    \vskip -.15in
    \centering%
    \subfloat[Concat]{%
        \centering%
        \scalebox{.6}{%
            \begin{tikzpicture}[shorten >=1pt,->,draw=black!50, node distance=.5cm]%
                \input{tikz/concat_net}%
            \end{tikzpicture}%
        }\label{fig:concatnet}%
    }%
    \subfloat[Context]{%
        \centering%
        \scalebox{.6}{%
            \begin{tikzpicture}[shorten >=1pt,->,draw=black!50, node distance=.5cm]%
                \input{tikz/context_net}%
            \end{tikzpicture}%
        }\label{fig:contextnet}%
    }%
    \caption{Schematic representations of considered architectures for learning when to make decisions, where $a_t$ is the action coming from a separate behaviour policy.%
    }%
    \label{fig:nets}%
    \vskip -.2in
\end{figure}%
When using deep function approximation for \temporal we have to carefully consider how we parameterize the skip policy.
Commonly, in deep RL we do not only deal with featurized states but also with image-based ones.
Depending on the state modality
we can consider different architectures:

\textbf{Concatenation}
The simplest parametrization of our skip-policy assumes that the state of the environment we are learning from is featurized, i.e., a state is a vector of individual informative features.
In this setting, the skip-policy network can take any architecture deemed appropriate for the environment, where the input is a concatenation of the original state $s_t$ and the chosen behaviour action $a_t$, i.e., $
s_t' = (s_t,a_t)$, see Figure~\ref{fig:concatnet}.
This allows the skip-policy network to directly learn features that take into account the chosen behaviour action.
However, note that this concatenation assumes that the state is already featurized.

\textbf{Contextualization}
In deep RL, we often have to learn to act directly from images.
In this case, concatenation is not trivially possible.
Instead we propose to use the behaviour action as context information further down-stream in the network.
Feature learning via convolutions can then progress as normal and the learned high-level features can be concatenated with the action $a_t$ and be used to learn the final skip-value, see Figure~\ref{fig:contextnet}.

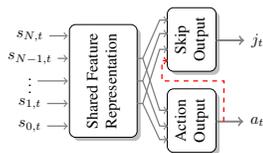
\begin{wrapfigure}{r}{0.22\textwidth}
    \centering
        \centering
        \scalebox{.6}{

            \begin{tikzpicture}[shorten >=1pt,->,draw=black!50, node distance=.5cm]
            \node (a) {$s_{0,t}$};
\node (b) [above of=a] {$s_{1,t}$};
\node (c) [above of=b] {$\vdots$};
\node (d) [above of=c] {$s_{N-1,t}$};
\node (e) [above of=d] {$s_{N,t}$};

\node (inBL) [right=.75cm of a.south] {};
\node (inUR) [right=2.25cm of d.north] {};
\node (inUUR) [right=2.25cm of e.north] {};
\path [activity] (inBL) rectangle (inUUR);
\node (inC) [rotate=90, align=center] at ($(inBL)!0.5!(inUUR)$) {Shared Feature\\Representation};

\draw[myarrow] (a) -- (inBL |- a);
\draw[myarrow] (b) -- (inBL |- b);
\draw[myarrow] (c) -- (inBL |- c);
\draw[myarrow] (d) -- (inBL |- d);
\draw[myarrow] (e) -- (inBL |- e);

\node (hiddenBL) [above right=-.3cm and 2.75cm of c] {};
\node (hiddenUR) [above right=.125cm and 1.5cm of inUUR] {};
\path [activity] (hiddenBL) rectangle (hiddenUR);
\node (hiddenC) [rotate=90, align=center] at ($(hiddenBL)!0.5!(hiddenUR)$) {Skip\\Output};

\node (hiddenAUR) [below=1.55cm of hiddenUR] {};
\node (hiddenABL) [below=1.55cm of hiddenBL] {};
\path [activity] (hiddenABL) rectangle (hiddenAUR);
\node (hiddenAC) [rotate=90, align=center] at ($(hiddenABL)!0.5!(hiddenAUR)$) {Action\\Output};

% From D to Skip
\draw[myarrow] (inUR |-d) -- ($(inUR |- d)+(0.125,0)$) -- ($(inUR |- e)+(.5,.35)$) -- ($(inUR |- e)+(.7125,.35)$);
% From C to Skip
\draw[myarrow] (inUR |-c) -- ($(inUR |- c)+(0.125,0)$) -- ($(inUR |- e)+(.5,0.05)$) -- ($(inUR |- e)+(.7125,.05)$);
% From B to Skip
\draw[myarrow] (inUR |-b) -- ($(inUR |- b)+(0.125,0)$) -- ($(inUR |- d)+(.5,.2)$) -- ($(inUR |- d)+(.7125,.2)$);

% From D to Act
\draw[myarrow] ($(inUR |- d)+(0.125,0)$) -- ($(inUR |- b)+(.5,-.15)$) -- ($(inUR |- b)+(.7125,-.15)$);
% From C to Act
\draw[myarrow] ($(inUR |- c)+(0.125,0)$) -- ($(inUR |- a)+(.5,0)$) -- ($(inUR |- a)+(.7125,0)$);
% From B to Act
\draw[myarrow] ($(inUR |- b)+(0.125,0)$) -- ($(inUR |- a)+(.5,-.3)$) -- ($(inUR |- a)+(.7125,-.3)$);

\node (cc) at ($(hiddenBL |- c)+(2,.9)$) {$j_{t}$};
\draw[myarrow,very thick] (hiddenUR |-cc) -- (cc);

\node (bb) at ($(hiddenABL |- c)+(2,-.9)$) {$a_{t}$};
\draw[myarrow,very thick] (hiddenAUR |-bb) -- (bb);

% From a_t+1 to j_t+1
\draw[myarrow,color=red,dashed] ($(bb) - (.75,0)$) -- ($(inUR |- d)+(1.925,-.5)$) -- ($(inUR |- d)+(.55,-.5)$) -- ($(inUR |- d)+(.55,-.05)$) -- ($(inUR |- d)+(.7125,-.05)$);
            \end{tikzpicture}
        }
    \caption{Architecture with shared feature representation for joint learning of when to make a decision and what action to take.}
    \label{fig:sharedtnet}
\end{wrapfigure}

\textbf{Shared~Weights}
Concatenation and contextualization learn individual policy networks for the behaviour and skip policies and do not share information between the two.
To achieve this we can instead share parts of the networks, e.g., the part of learning higher-level features from images (see Figure~\ref{fig:sharedtnet}).
This allows us to learn the two policy networks with potentially fewer weights than two
completely independently learned networks.
In the forward and backward passes,
only the shared feature representation with the corresponding output layers are active.
Similar to the contextualization, the output layers for the skip-values require the selected action, i.e. the argmax of the action outputs, as additional input.

%--------------------------------------------------------------------------------------------
%--------------------------------------------------------------------------------------------
%--------------------------------------------------------------------------------------------
\section{Experiments}
We evaluated \temporal with tabular as well as deep $\q$-functions. We first give results for the tabular case. All code, the appendix and experiment data including trained policies are available at \href{https://www.github.com/automl/TempoRL}{\nolinkurl{github.com/automl/TempoRL}}.
For details on the used hardware see Appendix~\ref{appendix:sec:compute}.

\subsection{Tabular TempoRL}
\begin{figure}[h
tbp]
    \vskip -0.15in
    \centering
    \subfloat[Cliff]{
        \centering
        \scalebox{.45}{
            \begin{tikzpicture}
                \input{tikz/cliff-grid}
                \draw[line width=0.75mm,my_magenta] (0,0) to[out=150,in=210, looseness=0.375] (0,1.5)  to[out=150,in=120, looseness=0.325]  (4.5,1.5) to[out=30,in=-30, looseness=0.375] (4.5,0);
                \draw[fill=my_green] (0,0) circle (0.125cm);
                \draw[fill=my_green] (0,0.5) circle (0.125cm);
                \draw[fill=my_green] (4.5,0.5) circle (0.125cm);
                \draw[fill=my_green] (0,1) circle (0.125cm);
                \draw[fill=my_green] (4.5,1) circle (0.125cm);
                \draw[line width=0.75mm,my_green,dashed] (0,0) -- (0,1.5) -- (4.5,1.5) -- (4.5,0);
                \foreach \x in {0,0.5,...,4.5} {
                    \draw[fill=my_green] (\x, 1.5) circle (0.125cm);
                }
                \draw[fill=my_magenta] (0,0) circle (0.0625cm);
                \draw[fill=my_magenta] (0,1.5) circle (0.0625cm);
                \draw[fill=my_magenta] (4.5,1.5) circle (0.0625cm);
                \node at (4.5,0) {\includegraphics[scale=0.125]{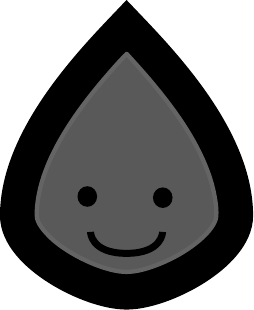}};
            \end{tikzpicture}
        }\label{fig:cliff}
    }%
    \subfloat[Bridge]{
        \centering
        \scalebox{.45}{
            \begin{tikzpicture}
                \input{tikz/bridge-grid}
                \draw[line width=0.75mm,my_magenta] (0,0) to[out=150,in=210, looseness=0.375] (0,1)  to[out=150,in=120, looseness=0.325]  (4.5,1) to[out=30,in=-30, looseness=0.375] (4.5,0);
                \draw[fill=my_green] (0,0) circle (0.125cm);
                \draw[fill=my_green] (0,0.5) circle (0.125cm);
                \draw[fill=my_green] (4.5,0.5) circle (0.125cm);
                \draw[line width=0.75mm,my_green,dashed] (0,0) -- (0,1) -- (4.5,1) -- (4.5,0);
                \foreach \x in {0,0.5,...,4.5} {
                    \draw[fill=my_green] (\x, 1) circle (0.125cm);
                }
                \draw[fill=my_magenta] (0,0) circle (0.0625cm);
                \draw[fill=my_magenta] (0,1) circle (0.0625cm);
                \draw[fill=my_magenta] (4.5,1) circle (0.0625cm);
                \node at (4.5,0) {\includegraphics[scale=0.125]{crc_images/agent.pdf}};
            \end{tikzpicture}
        }\label{fig:bridge}
    }%
    \subfloat[ZigZag]{
        \centering
        \scalebox{.45}{
            \begin{tikzpicture}
                \input{tikz/zigzag-grid}
                \draw[line width=0.75mm,my_magenta] (0,0) 
                to[out=150,in=210, looseness=0.375] (0,2)  
                to[out=150,in=120, looseness=0.325]  (2.5,2) 
                to[out=30,in=-30, looseness=0.375] (2.5,.5)
                to[out=-150,in=-120, looseness=0.375] (4.5,.5)
                to[out=-150,in=-210, looseness=0.375] (4.5,2.5);
                \draw[fill=my_green] (0,0) circle (0.125cm);
                \draw[fill=my_green] (0,0.5) circle (0.125cm);
                \draw[fill=my_green] (0,1) circle (0.125cm);
                \draw[fill=my_green] (0,1.5) circle (0.125cm);
                \draw[line width=0.75mm,my_green,dashed] (0,0) -- (0,2) -- (2.5,2) -- (2.5,.5) -- (4.5,.5) -- (4.5,2.5);
                \foreach \x in {0,0.5,...,2.5} {
                    \draw[fill=my_green] (\x, 2) circle (0.125cm);
                }
                \foreach \x in {2.5,3,...,4.5} {
                    \draw[fill=my_green] (\x, .5) circle (0.125cm);
                }
                \foreach \y in {1,1.5,2,2.5} {
                    \draw[fill=my_green] (4.5, \y) circle (0.125cm);
                }
                \foreach \y in {1,1.5} {
                    \draw[fill=my_green] (2.5, \y) circle (0.125cm);
                }
                \draw[fill=my_magenta] (0,0) circle (0.0625cm);
                \draw[fill=my_magenta] (0,2) circle (0.0625cm);
                \draw[fill=my_magenta] (2.5,2) circle (0.0625cm);
                \draw[fill=my_magenta] (2.5,0.5) circle (0.0625cm);
                \draw[fill=my_magenta] (4.5,0.5) circle (0.0625cm);
                \draw[fill=my_magenta] (4.5,2.5) circle (0.0625cm);
                \node at (4.5,2.5) {\includegraphics[scale=0.125, angle=180]{crc_images/agent.pdf}};
            \end{tikzpicture}
        }\label{fig:zigzag}
    }
    \caption{$6\times10$ Grid Worlds. Agents have to reach a fixed {\color{my_blue} goal state} from a fixed {\color{my_gray} start state}. Dots represent decision steps of {\color{my_green} vanilla}  and {\color{my_magenta} \temporal} \qlearning policies.}
    \label{fig:grids}
    \vskip -.15in
\end{figure}

In this subsection, we describe experiments for a tabular \qlearning implementation that we evaluated on various grid-worlds with sparse rewards (see Figure~\ref{fig:grids}).
We first evaluate our approach on the \textit{cliff} environment (see Figure~\ref{fig:cliff}) before evaluating the influence of the exploration schedule on both vanilla and \temporal \qlearning, which we refer to as {\color{my_green}$\q$} and {\color{my_magenta}$t$-$\q$}, respectively.
{
\begin{figure*}[tbp]
    \vskip -0.1in
    \centering
    \subfloat[Cliff -- Reward per Episode]{
    \includegraphics[width=.34\textwidth]{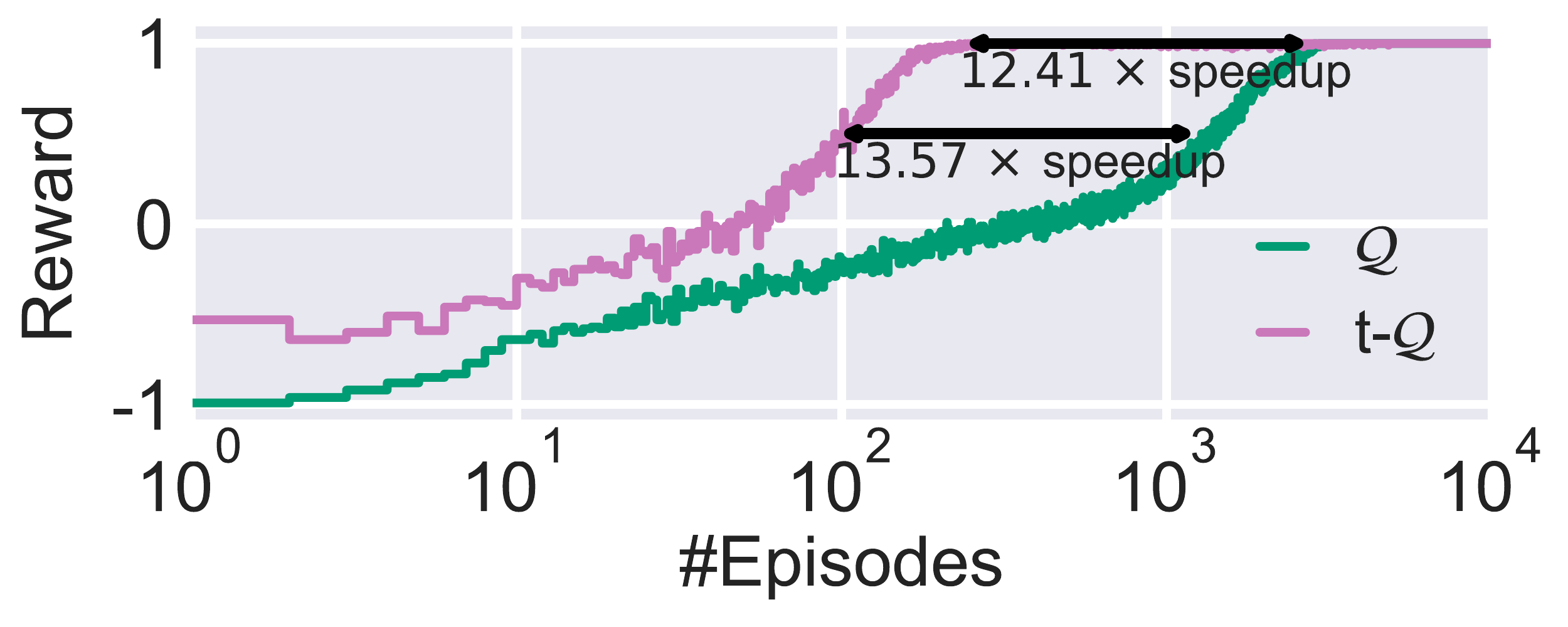}\label{fig:cliff_perf}}
    \subfloat[Cliff -- Steps per Episode]{
    \includegraphics[width=.34\textwidth]{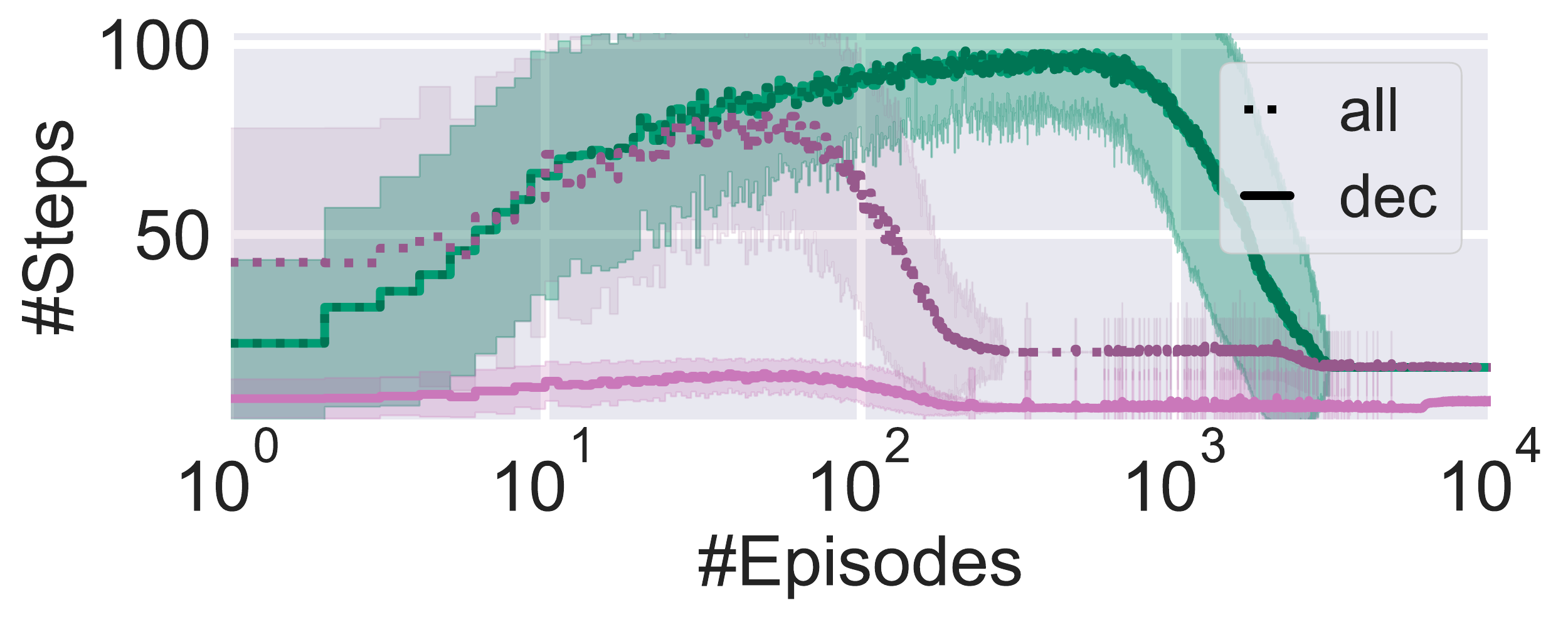}\label{fig:cliff_lens}}
    \subfloat[Temporal Exploration]{\includegraphics[width=0.275\textwidth]{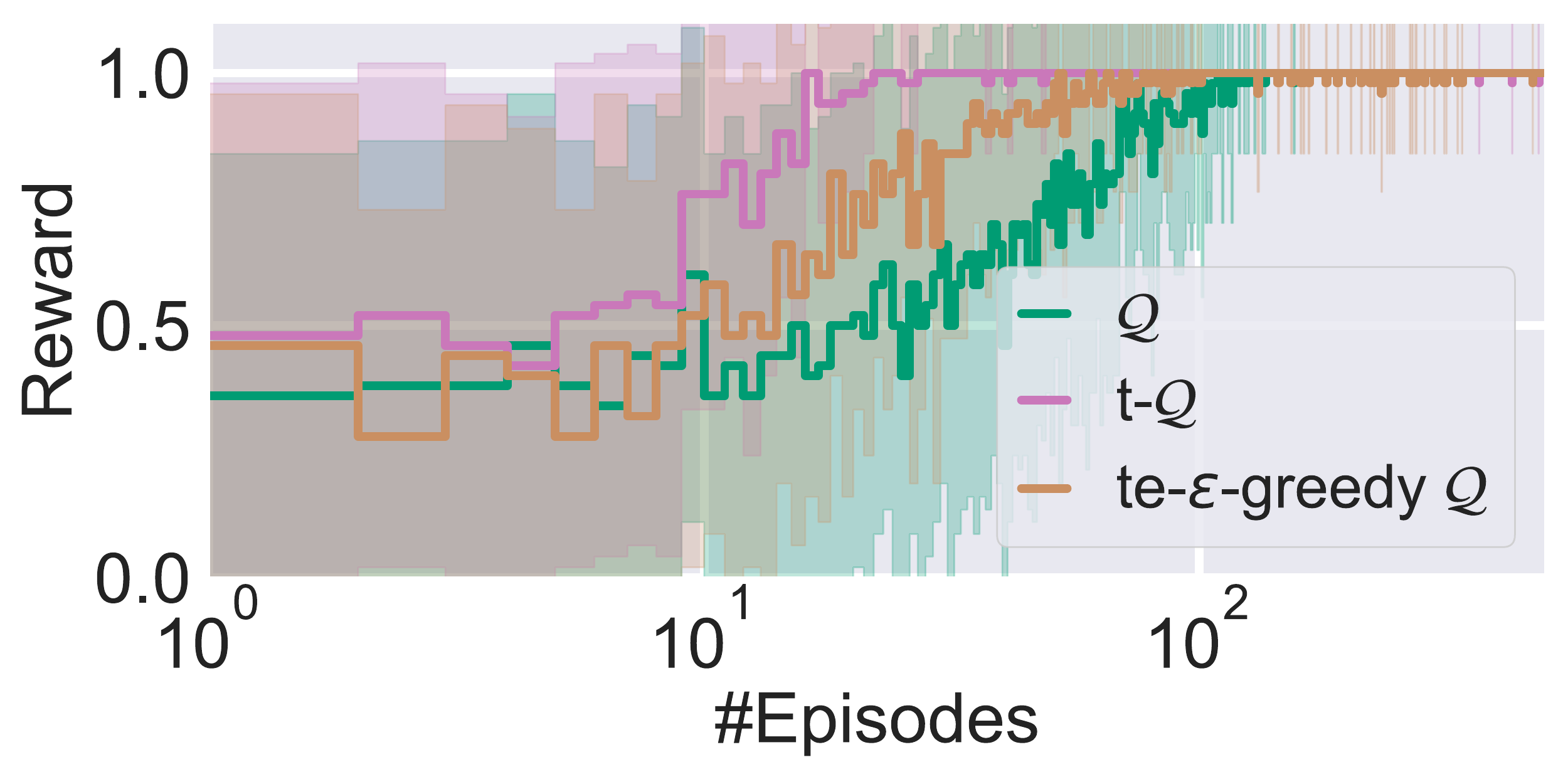}\label{fig:dabney}}
    \vskip -0.1in
    \caption{
    Evaluation performance of tabular \qlearning agents over $100$ random seeds. \protect\subref{fig:cliff_perf} \& \protect\subref{fig:cliff_lens}: The agents were trained with a linearly-decaying $\epsilon$-greedy policy on the cliff environment.
    \protect\subref{fig:cliff_perf} Achieved reward. \protect\subref{fig:cliff_lens} Length of executed policy ($\cdots$) and number of decisions (---) made by the policies.
    \protect\subref{fig:dabney} Comparison to temporally extended $\epsilon$-greedy exploration (te-$\epsilon$-greedy $\mathcal{Q}$ in plot) on a $23\times23$ Gridworld~\cite{dabney-arxiv20}.
    {\color{my_magenta} t-$\mathcal{Q}$} is our proposed \temporal agent. The lines/shaded area represent the mean/standard deviation.
    }
    \label{fig:cliff_both}
    \vskip -0.1in
\end{figure*}
}

\textbf{Gridworlds} All considered environments (see Figure~\ref{fig:grids}) are discrete, deterministic, have sparse rewards and have size $6\times10$. Falling off a cliff results in a negative reward~($-1$) and reaching a goal state results in a positive reward ($+1$). For a more detailed description of the gridworld environments we refer to Appendix~\ref{appendix:sec:grids}.

For this experiment, we limit our \temporal agent to a maximum skip length of $J=7$; thus, a learned optimal policy requires $4$ decision points instead of $3$. 
For evaluations using larger skips we refer to Appendix~\ref{appendix:skip-length}.
Note that increasing the skip-length improves \temporal up to some point, at which it has too many irrelevant skip-actions at its disposal which slightly decreases the performance.
We compare the learning speed, in terms of training policies, of {\color{my_magenta} our approach} to a {\color{my_green}vanilla \qlearning agent}. Both methods are trained for $10\,000$ episodes using the same $\epsilon$-greedy strategy, where $\epsilon$ is linearly decayed from $1.0$ to $0.0$ over all episodes.

Figure~\ref{fig:cliff_perf} depicts the evaluation performance of both methods. \temporal is $13.6\times$ faster than its vanilla counterpart to reach a reward of $0.5$, and $12.4\times$ faster to reach a reward of $1.0$ (i.e., always reach the goal).
Figure~\ref{fig:cliff_lens} shows the number of required steps in the environment, as well as the number of decision steps.
\temporal is capable of finding a policy that reaches the goal much faster than vanilla \qlearning while requiring far fewer decision steps.
Furthermore, \temporal recovers the optimal policy quicker than vanilla \qlearning.
Lastly we can observe that after having trained for $\approx6\,000$ episodes, \temporal starts to increase the number of decision points.
This can be attributed to skip values of an action having converged to the same value and our implementation selecting a random skip as tie-breaker.%
\begin{table}[tbp!]%
    \vskip -0.1in%
    \centering%
    \caption{Normalized AUC for reward and average number of decision steps. Both agents are trained with the same $\epsilon$ schedule.}
    \label{tab:envs}
    \vskip 0.1in
    \subfloat[linearly decaying $\epsilon$-schedule]{
    \begin{tabular}{lrrrrrr}
    \toprule
         \multicolumn{1}{c}{} & \multicolumn{2}{c}{Cliff} & \multicolumn{2}{c}{Bridge} & \multicolumn{2}{c}{ZigZag}\\
         \cmidrule(lr){2-3} \cmidrule(lr){4-5} \cmidrule(lr){6-7}
         & {\color{my_green}$\q$} & {\color{my_magenta}$t\mhyphen\q$} & {\color{my_green}$\q$} & {\color{my_magenta}$t\mhyphen\q$} & {\color{my_green}$\q$} & {\color{my_magenta}$t\mhyphen\q$}\\
        \midrule
        \scalebox{.9}{Reward$_{\mathrm{AUC}}$}  & \scalebox{.9}{$0.92$} & \scalebox{.9}{$\bm{0.99}$} & \scalebox{.9}{$0.75$} & \scalebox{.9}{$\bm{0.97}$} & \scalebox{.9}{$0.57$} & \scalebox{.9}{$\bm{0.92}$}\\
        \scalebox{.9}{Decisions} & \scalebox{.9}{$27.9$} & \scalebox{.9}{$\bm{5.2}$} & \scalebox{.9}{$49.5$} & \scalebox{.9}{$\bm{5.0}$} & \scalebox{.9}{$83.6$} & \scalebox{.9}{$\bm{7.9}$}\\
    \bottomrule
    \end{tabular}\label{tab:envs_linear}
    }
    
    \subfloat[logarithmically decaying $\epsilon$-schedule]{
    \begin{tabular}{lrrrrrr}
    \toprule
        \scalebox{.9}{Reward$_{\mathrm{AUC}}$}   & \scalebox{.9}{$0.96$} & \scalebox{.9}{$\bm{0.99}$} & \scalebox{.9}{$0.94$} & \scalebox{.9}{$\bm{0.98}$} & \scalebox{.9}{$0.90$} & \scalebox{.9}{$\bm{0.96}$}\\
        \scalebox{.9}{Decisions} & \scalebox{.9}{$21.7$} & \scalebox{.9}{$\bm{4.9}$} & \scalebox{.9}{$21.4$} & \scalebox{.9}{$\bm{5.3}$} & \scalebox{.9}{$35.6$} & \scalebox{.9}{$\bm{6.9}$}\\
    \bottomrule
    \end{tabular}\label{tab:envs_log}
    }
    
    \subfloat[constant $\epsilon=0.1$]{
    \begin{tabular}{lrrrrrr}
    \toprule
        \scalebox{.9}{Reward$_{\mathrm{AUC}}$}    & \scalebox{.9}{$\bm{0.99}$} & \scalebox{.9}{$\bm{0.99}$} & \scalebox{.9}{$0.98$} & \scalebox{.9}{$\bm{0.99}$} & \scalebox{.9}{$0.95$} & \scalebox{.9}{$\bm{0.99}$}\\
        \scalebox{.9}{Decisions} & \scalebox{.9}{$17.1$} & \scalebox{.9}{$\bm{5.1}$} & \scalebox{.9}{$14.7$} & \scalebox{.9}{$\bm{5.2}$} & \scalebox{.9}{$27.6$} & \scalebox{.9}{$\bm{7.1}$}\\
    \bottomrule
    \end{tabular}\label{tab:envs_const}
    }
    \label{tab:tabular_all}
    \vskip -0.35in
\end{table}%

Table~\ref{tab:tabular_all} summarizes the results on all environments in terms of normalized area under the reward curve and number of decisions for three different $\epsilon$-greedy schedules.
A reward AUC value closer to $1.0$ indicates that the agent was capable of learning to reach the goal quickly.
A lower number of decisions is better as fewer decisions were required to reach the goal, making a policy easier to learn.
In view of both metrics, \temporal readily outperforms the vanilla agent, learning much faster and requiring far fewer decisions.

\textbf{Sensitivity to Exploration} As the used exploration mechanism can have a dramatic impact on agent performance we evaluated the agents for three commonly used $\epsilon$-greedy exploration schedules.
In the cases of linearly and logarithmically decaying schedules, we decay $\epsilon$ over all $10\,000$ training episodes, starting from $1.0$ and decaying it to $0$ or $10^{-5}$, respectively.
In the constant case, we set $\epsilon=0.1$.

As shown in Table~\ref{tab:envs}, maybe not surprisingly, too much (linear) and too little (log) exploration are both detrimental to the agent's performance.
However, \temporal performs quite robustly even using suboptimal exploration strategies.
\temporal outperforms its vanilla counterpart in all cases, showing the effectiveness of our proposed method.

\textbf{Guiding Exploration}
To demonstrate \temporal not only benefits through better exploration but also learning \textit{when} to act, we use the $23\times23$ Gridworld and agent hyperparameters as introduced by \citet{dabney-arxiv20}.

An agent starts in the top center and has to find a goal further down and to the left only getting a reward for reaching the goal within $1000$ steps.
Temporally-extended exploration \citep[te-$\epsilon$-greedy \Q-learning;][]{dabney-arxiv20} is able to cover a space much better than 1-step exploration.
However,
it falls short in guiding the agent back to high reward areas.
\temporal enables an agent to quickly find a successful policy that reach a goal while exploring around such a policy.
Figure~\ref{fig:dabney} shows \temporal reliably reaches the goal after $\approx30$ episodes.
An agent using temporally-extended epsilon greedy exploration does not reliably reach the goal in this time-frame and on average requires twice as many steps.

\subsection{Deep TempoRL}
In this section, we describe experiments for agents using deep function approximation implemented with PyTorch~\cite{pytorch-neurips19a} in version 1.4.0. We begin with experiments on featurized environments before evaluating on environments with image states.
We evaluate \temporal for DQN with different architectures for the skip $\q$-function.
We compare against \emph{dynamic action repetition} \citep[DAR;][]{lakshminarayanan-aaai17} for the DQN experiments and against
\emph{Fine grained action repetition} \citep[FiGAR][]{sharma-iclr17} for experiments with DDPG.\footnote{Neither DAR, nor FiGAR are publicly available and thus we used our own reimplementation available at \href{https://www.github.com/automl/TempoRL}{\nolinkurl{github.com/automl/TempoRL}}.}
%--------------------------------------------------------------------------------------------
%--------------------------------------------------------------------------------------------
%--------------------------------------------------------------------------------------------

\begin{table*}[htb]%
    \centering%
    \caption{Average normalized reward AUC for DDPG agents on Pendulum-v0. t-DDPG and FiGAR are evaluated over different maximal skip-lengths for $15$ seeds. Corresponding learning curves are given in Appendix~\ref{appendix:sec:ddpg}}
    \label{tab:auc_deep_featurized_ddpg}
    \vskip 0.1in
    \begin{tabular}{ccccccccccccccc}
        \toprule
        {} & \multicolumn{7}{c}{\color{my_tddpg_blue}t-DDPG} & \multicolumn{7}{c}{\color{my_figar_gray}FiGAR}\\
        \cmidrule(lr){1-1} \cmidrule(lr){2-8} \cmidrule(lr){9-15}
        {\color{my_ddpg_orange}DDPG} & 
        $2$ &  $4$ &  $6$ &  $8$ &  $10$ &
        $14$ &
        $20$ &
        $2$ &  $4$ &  $6$ &  $8$ &  $10$ &
        $14$ &
        $20$ \\
        \cmidrule(lr){1-1} \cmidrule(lr){2-8} \cmidrule(lr){9-15}
        \scalebox{.9}{$\bm{\mathit{0.92}}$} &
        \scalebox{.9}{$0.89$} & \scalebox{.9}{$0.89$} &
        \scalebox{.9}{$\mathit{0.90}$} & \scalebox{.9}{$0.89$} & 
        \scalebox{.9}{$0.89$} &
        \scalebox{.9}{$0.89$} &
        \scalebox{.9}{$0.88$} &
        \scalebox{.9}{$\mathit{0.76}$} & \scalebox{.9}{$0.57$} &
        \scalebox{.9}{$\mathit{0.39}$} & \scalebox{.9}{$0.31$} & 
        \scalebox{.9}{$0.28$} &
        \scalebox{.9}{$0.25$} &
        \scalebox{.9}{$0.24$}
        \\
        \bottomrule
    \end{tabular}
    \vskip -0.15in
\end{table*}
\subsubsection{Adversarial Environment -- DDPG}\label{sec:experiments_featurized_ddpg}
\textbf{Setup}
We chose to first evaluate on OpenAI gyms~\cite{brockman-arxiv16a} Pendulum-v0 as it is an adversarial setting where high action repetition is nearly guaranteed to overshoot the balancing point.
Thus, agents using action repetition that make mistakes during training will have to spend additional time learning \emph{when} it is necessary to be reactive; a challenge vanilla agents are not faced with.
We trained all DDPG agents~\cite{lillicrap-iclr16} for a total of $3\times10^4$ training steps and evaluated the agents every $250$ training steps.
The first $10^3$ steps follow a uniform random policy to generate the initial experience.
We used Adam~\cite{kingma-iclr15} with PyTorchs default settings.

\textbf{Agents} 
All actor and critic networks of all {\color{my_ddpg_orange}DDPG} agents consist of two hidden layers with $400$ and $300$ hidden units respectively.
Following \cite{sharma-iclr17}, {\color{my_figar_gray}FiGAR} introduces a second actor network that shares the input layer with the original actor network.
The output layer is a softmax layer with $J$ outputs, representing the probability of repeating the action for $j\in\left\{1,\ldots,J\right\}$ time-steps.
Both actor outputs are jointly input to the critic and gradients are directly propagated from the critic through both actors.
\temporal DDPG (which we refer to as {\color{my_tddpg_blue}t-DDPG} in the following) uses the concatenation architecture which takes the state with the action output of the DDPG actor as input and makes use of the critic's $\q$-function when learning the skip $\q$-function.
We evaluate t-DDPG and FiGAR on a grid of maximal skip lengths of $\left\{2, 4, 6,\ldots, 20\right\}$.
See Appendix~\ref{appendix:sec:ddpg} for implementation details and all used hyperparameters.

\textbf{Pendulum} Table~\ref{tab:auc_deep_featurized_ddpg} confirms that agents using action repetition indeed are slower in learning successful policies, as reflected by the normalized reward AUC.
As FiGAR does not directly inform the skip policy about the chosen repetition value or vica versa, the agent tends to struggle quite a lot in this environment already with only two possible skip-values and is not capable of handling larger maximal skip values.
In contrast to that, t-DDPG only slightly lags behind vanilla DDPG and readily adapts to larger skip lengths, by quickly learning to ignore irrelevant skip-values.
Further, due to making use of n-step learning, t-DDPG starts out very conservative as large skip values appear to lead to larger negative rewards in the beginning.
With more experience however, t-DDPG learns \emph{when} switching between actions becomes advantageous, thereby approximately halving the required decisions (see Appendix~\ref{appendix:sec:ddpg}).

%--------------------------------------------------------------------------------------------
%--------------------------------------------------------------------------------------------
%--------------------------------------------------------------------------------------------

\begin{table*}[htb!]%
    \centering%
    \caption{Average normalized reward AUC for different \temporal architectures and maximal skip-lengths over $50$ seeds. All agents are trained with the same $\epsilon$ schedule. Bold faced values give the overall best AUC and cursive values the best per architecture.}
    \label{tab:auc_deep_featurized_arch_comp}
    \vskip 0.1in
    \subfloat[MountainCar-v0]{\label{tab:mcauc_archs}
    \begin{tabular}{lccccc}
        \toprule
        \multicolumn{1}{c}{\scalebox{.9}{Max Skip}} & $2$ &  $4$ &  $6$ &  $8$ &  $10$\\
        \cmidrule(lr){1-1} \cmidrule(lr){2-6}
        \scalebox{.9}{concat} &  
        \scalebox{.9}{$0.469$} & \scalebox{.9}{$0.523$} & \scalebox{.9}{$0.602$} & \scalebox{.9}{$0.626$} & \scalebox{.9}{$\mathit{0.630}$}\\
        \scalebox{.9}{context} &   
        \scalebox{.9}{$0.429$} & \scalebox{.9}{$0.540$} & \scalebox{.9}{$0.601$} & \scalebox{.9}{$0.608$} & \scalebox{.9}{$\mathit{0.620}$}\\  
        \scalebox{.9}{shared} &  
        \scalebox{.9}{$0.440$} & \scalebox{.9}{$0.464$} & \scalebox{.9}{$0.592$} & \scalebox{.9}{$0.561$} & \scalebox{.9}{$\bm{\mathit{0.644}}$}\\
        \bottomrule
    \end{tabular}
    }%
    \subfloat[LunarLander-v2]{\label{tab:llauc_archs}
    \begin{tabular}{lccccc}
        \toprule
        \multicolumn{1}{c}{\scalebox{.9}{Max Skip}} & $2$ &  $4$ &  $6$ &  $8$ &  $10$\\
        \cmidrule(lr){1-1} \cmidrule(lr){2-6}
        \scalebox{.9}{concat} &  
        \scalebox{.9}{$0.855$} & \scalebox{.9}{$\bm{\mathit{0.878}}$} & \scalebox{.9}{$0.868$} & \scalebox{.9}{$0.862$} & \scalebox{.9}{$0.830$}\\
        \scalebox{.9}{context} &   
        \scalebox{.9}{$0.858$} & \scalebox{.9}{$\mathit{0.876}$} & \scalebox{.9}{$0.871$} & \scalebox{.9}{$0.859$} & \scalebox{.9}{$0.837$}\\  
        \scalebox{.9}{shared} &  
        \scalebox{.9}{$\mathit{0.851}$} & \scalebox{.9}{$0.837$} & \scalebox{.9}{$0.803$} & \scalebox{.9}{$0.769$} & \scalebox{.9}{$0.696$}\\
        \bottomrule
    \end{tabular}
    }
    \vskip -0.2in
\end{table*}
\subsubsection{Featurzied Environments -- DQN}\label{sec:experiments_featurized}
\textbf{Setup} 
We trained all agents for a total of $10^6$ training steps using a constant $\epsilon$-greedy exploration schedule with $\epsilon$ set to $0.1$. We evaluated all agents every $200$ training steps.
We used the Adam with a learning rate of $10^{-3}$ and default parameters as given in PyTorch v1.4.0.
For increased learning stability, we implemented all agents using double deep $\q$ networks~\cite{hasselt-aaai16}.
All agents used a replay buffer with size $10^6$ and a discount factor $\gamma$ of $0.99$.
The \temporal agents used an additional replay buffer of size $10^6$ to store observed skip-transitions.
We used the MountainCar-v0 and LunarLander-v2 environments.
See Appendix~\ref{appendix:sec:feats} for a detailed description of the environments.

\textbf{Agents}
The basic {\color{my_green} DQN} architecture consists of $3$ layers, each with $50$ hidden units and \emph{ReLu} activation functions.
The {\color{my_gold}DAR} baseline used the same architecture as the DQN agent but duplicated the output heads twice, each of which is associated with specific repetition values allowing for fine and coarse control.
We evaluated possible coarse control values on the grid $\left\{2, 4, 6, 8, 10\right\}$, keeping the fine-control value fixed to $1$ to allow for actions at every-time step.

For {\color{my_magenta}\temporal} agents not using weight sharing we used the same DQN architecture for both $\q$-functions.
The concatenation architecture used an additional input unit whereas the context architecture added the behaviour action as context at the third layer after using $10$ additional hidden units to process the behaviour action.
An agent using a weight-sharing architecture shared the first two layers of the DQN architecture and used the third layer of the DQN architecture to compute the behaviour $\q$-values.
The skip-output used $10$ hidden units to process the behaviour action and processed this output together with the hidden state of the 2nd layer in a 3rd layer with $60$ hidden units.
We refer to a DQN using \temporal as {\color{my_magenta}t-DQN} in the following.

\textbf{Influence of the Skip-Architecture}
We begin by evaluating the influence of architecture choice on our {\color{my_magenta}t-DQN} on both environments, before giving a more in-depth analysis on the learning behaviour in the individual environments.
To this end, we report the normalized reward AUC for all three proposed architectures and different maximal skip-lengths, see Table~\ref{tab:auc_deep_featurized_arch_comp}.
Both the concat and context architectures behave similarly on both environments, which is to be expected as they differ very little in setup.
Both architectures have an increase in AUC before reaching the best maximal skip-length for the respective environment.
The shared architecture, mostly conceptualized for image-based environments, however shows more drastic reactions to choice of $J$, leading to the best result in the first and to the worst result in the other environment.

\begin{table}[tb]%
    \centering%
    \caption{Average normalized reward AUC for maximal skip-length of $10$ for MountainCar-v0 and $4$ for LunarLander-v2 over $50$ seeds. All agents are trained with the same $\epsilon$ schedule. We show varying $^{max}_{min}$ repetitions for {\color{my_gold}DAR} and the best {\color{my_magenta}t-DQN} architecture (see Table~\ref{tab:auc_deep_featurized_arch_comp}). Bold faced values give the overall best AUC and cursive values the best per method which are plotted in Figure~\ref{fig:comp_featurized}.}
    \label{tab:auc_deep_featurized}
    \vskip 0.1in
    \subfloat[MountainCar-v0]{\label{tab:mcauc}
    \begin{tabular}{lccccccccc}
        \toprule
        {} & {} & \multicolumn{5}{c}{\color{my_gold}DAR}\\
        \cmidrule(lr){1-1} \cmidrule(lr){2-2} \cmidrule(lr){3-7}
        {\color{my_green}DQN} &   {\color{my_magenta}t-DQN} &  $_{1}^{2}$ &  $_{1}^{4}$ &  $_{1}^{6}$ &  $_{1}^{8}$ &  $_{1}^{10}$\\
        \cmidrule(lr){1-1} \cmidrule(lr){2-2} \cmidrule(lr){3-7}
        \scalebox{.9}{$\mathit{0.50}$} &  \scalebox{.9}{$\bm{\mathit{0.64}}$} & \scalebox{.9}{$0.43$} & \scalebox{.9}{$0.45$} & \scalebox{.9}{$\mathit{0.60}$} & \scalebox{.9}{$0.56$} &   \scalebox{.9}{$0.56$}\\
        \bottomrule
    \end{tabular}
    }
    
    \subfloat[LunarLander-v2]{\label{tab:llauc}
    \begin{tabular}{lccccccccc}
        \toprule
        \scalebox{.9}{$\mathit{0.83}$} &  \scalebox{.9}{$\bm{\mathit{0.88}}$} & \scalebox{.9}{$0.84$} & \scalebox{.9}{$\mathit{0.85}$} & \scalebox{.9}{$0.81$} & \scalebox{.9}{$0.72$} &   \scalebox{.9}{$0.60$}\\
        \bottomrule
    \end{tabular}
    }
    \vskip -0.35in
\end{table}

\begin{figure*}[tb]%
    \captionsetup[subfigure]{labelformat=empty}
    \centering%
    \subfloat{\centering%
    \includegraphics[width=.42\textwidth]{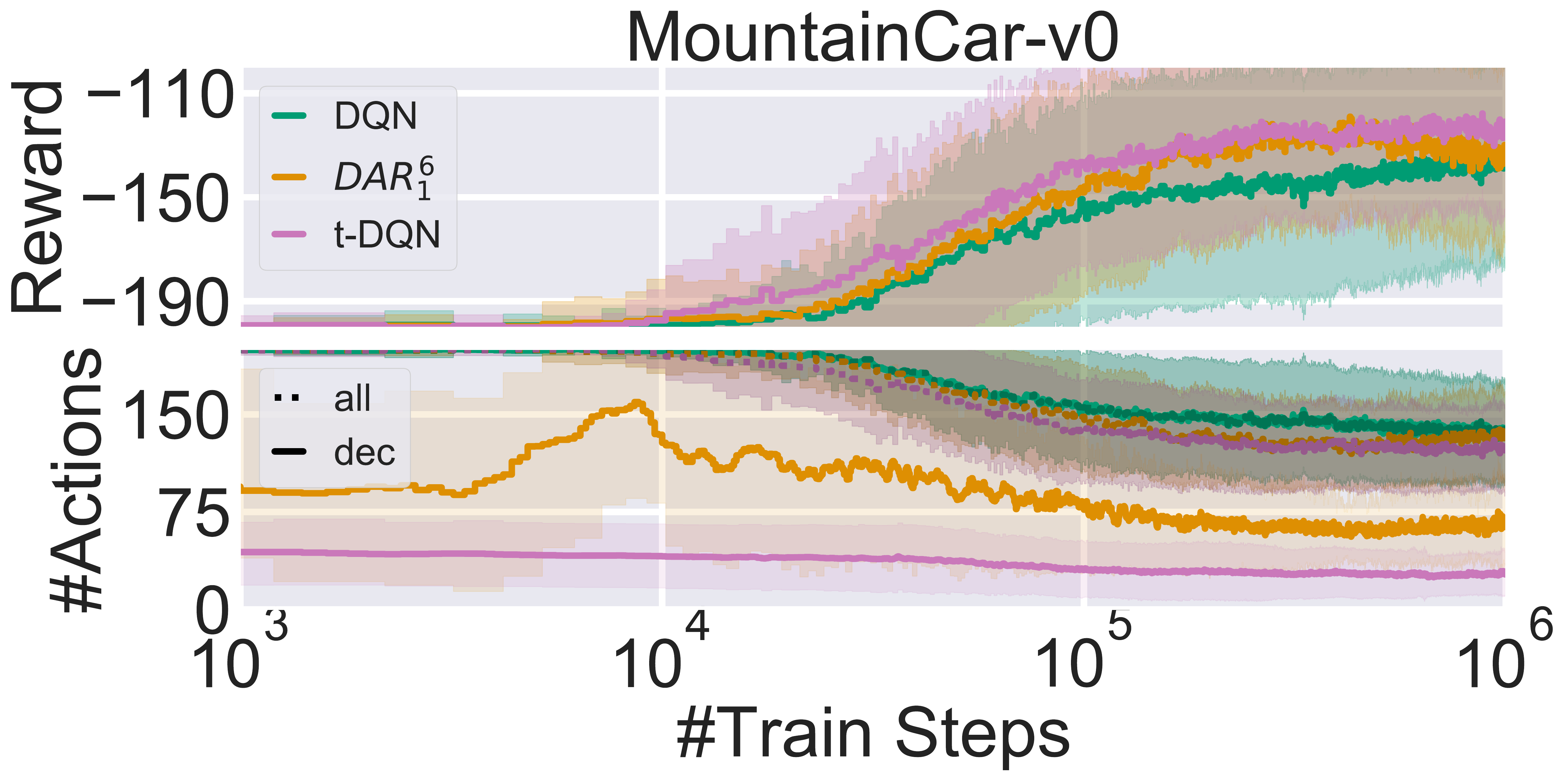}\label{fig:mountain_perf}}%
    \hspace{1cm}
    \subfloat{\centering%
    \includegraphics[width=.42\textwidth]{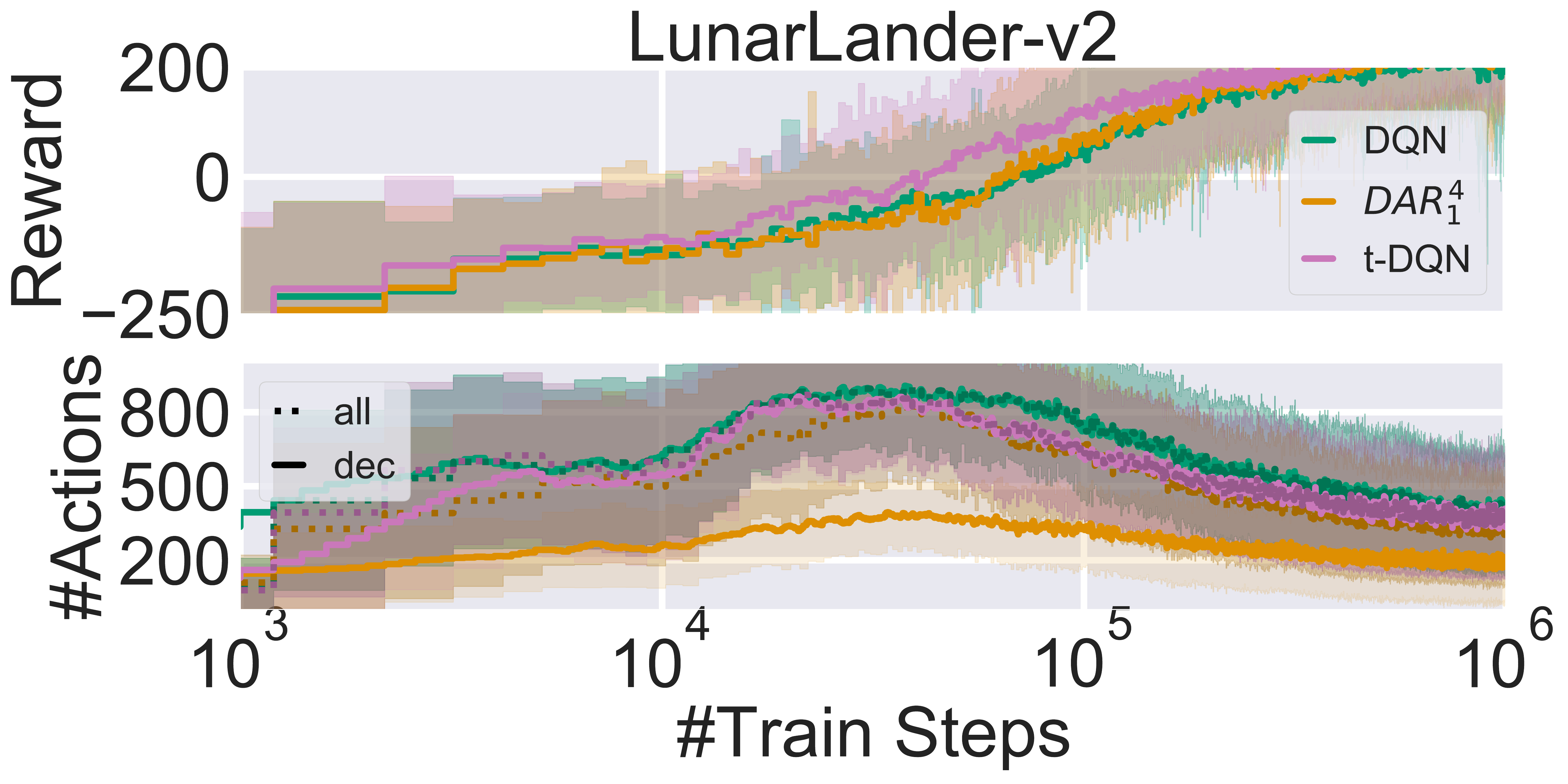}\label{fig:lunar_perf}}%
    \caption{%
    Evaluation performance of deep \qlearning agents on MountainCar-v0 and LunarLander-v2. Solid lines give the mean and the shaded area the standard deviation over $50$ random seeds. The sub- and superscripts of {\color{my_gold} DAR} give the best found fine and coarse repetition values respectively. {\color{my_magenta}t-DQN} is our proposed method using the best architecture as reported in Table~\ref{tab:auc_deep_featurized_arch_comp}. %
    (top) Achieved rewards. (bottom) Length of executed policy ($\cdots$) and number of decisions (---) made by the policies.}%
    \label{fig:comp_featurized}%
    \vskip -.15in
\end{figure*}

\textbf{MountainCar} 
Tables \ref{tab:mcauc_archs} \& \ref{tab:mcauc} depict the performance of the agents for different maximal skip lengths and Figure~\ref{fig:mountain_perf} shows the learning curves of the best \temporal architecture as well as the best found DAR agent.
On MountainCar the DQN baseline struggles in learning a successful policy, resulting in a small AUC of $0.50$ compared to the best result of t-DQN of $0.64$.
Furthermore, a well tuned DAR baseline, carefully trading off fine control and coarse control results in an AUC of $0.593$.
Figure~\ref{fig:mountain_perf} shows that DAR learns to trade off both coarse and fine-control.
However, as DAR does not know that two output heads correspond to the same action, with different repetition values, DARs reward begins to drop in the end as it learns to overly rely on coarse control.
During the whole training procedure the best t-DQN agent and the best DAR agent result in policies that require far fewer decisions, with t-DQN requiring only $\approx50$ decisions per episode reducing the number of decisions by a factor of $\approx3$ compared to vanilla DQN.

\textbf{LunarLander}
For such a dense reward there is only a small improvement for t-DQN and a properly tuned DAR agent.
Again the t-DQN agent performs best, achieving a slightly higher AUC of $0.88$ than the best tuned DAR agent ($0.85$), see Tables \ref{tab:llauc_archs} \& \ref{tab:llauc}.
Further, Figure~\ref{fig:lunar_perf} shows that, in this setting, t-DQN agents quickly learn to be very reactive, acting nearly at every time-step.
Again, DAR can not learn that some output heads apply the same behaviour action for multiple time-steps, preferring coarse over fine control.

%--------------------------------------------------------------------------------------------
%--------------------------------------------------------------------------------------------
%--------------------------------------------------------------------------------------------
\begin{figure*}[tb]
    \captionsetup[subfigure]{labelformat=empty}
    \centering
    \subfloat{\centering
    \includegraphics[width=.32\textwidth]{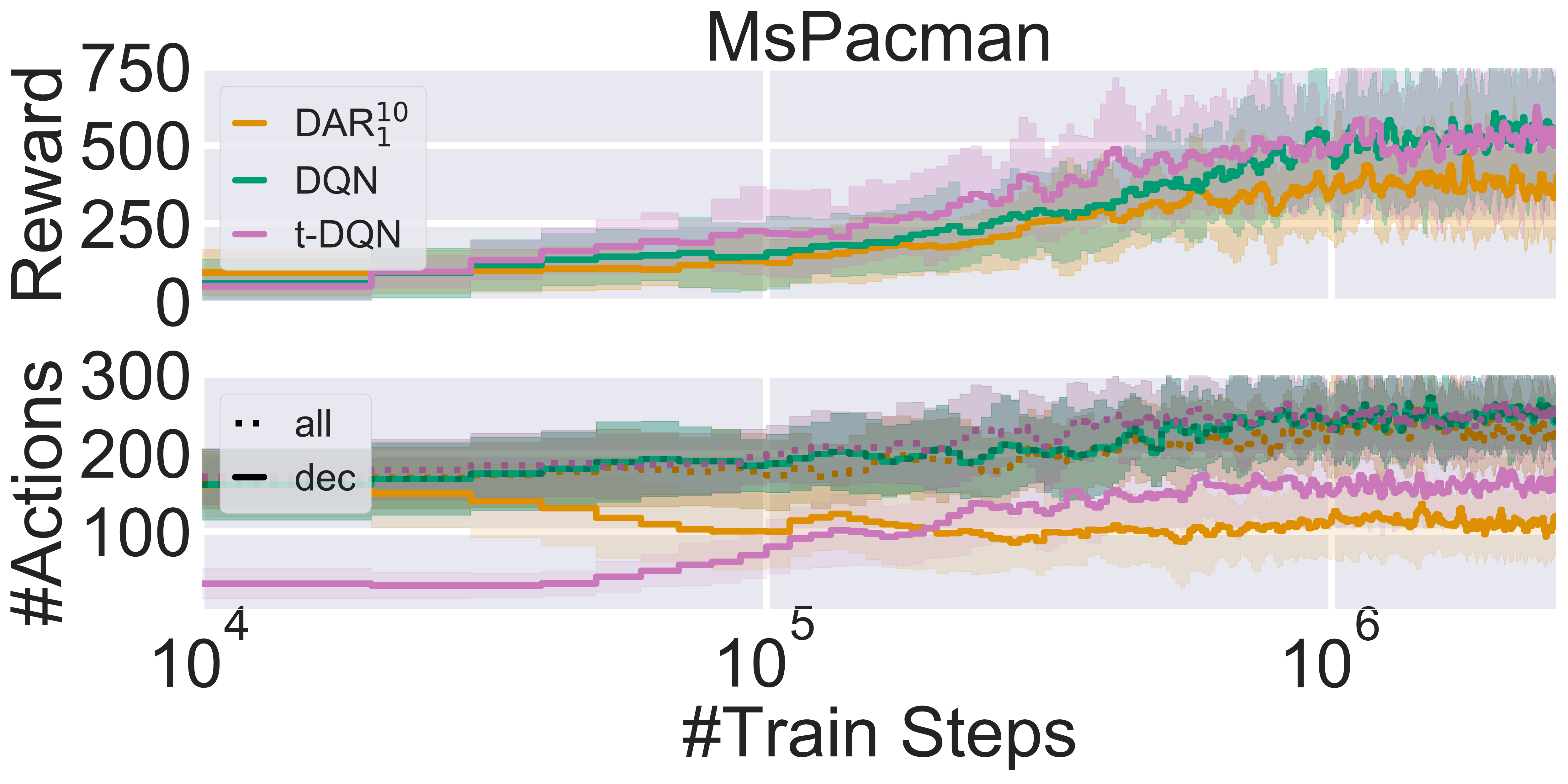}\label{fig:mspc_perf}}
    \subfloat{\centering
    \includegraphics[width=.32\textwidth]{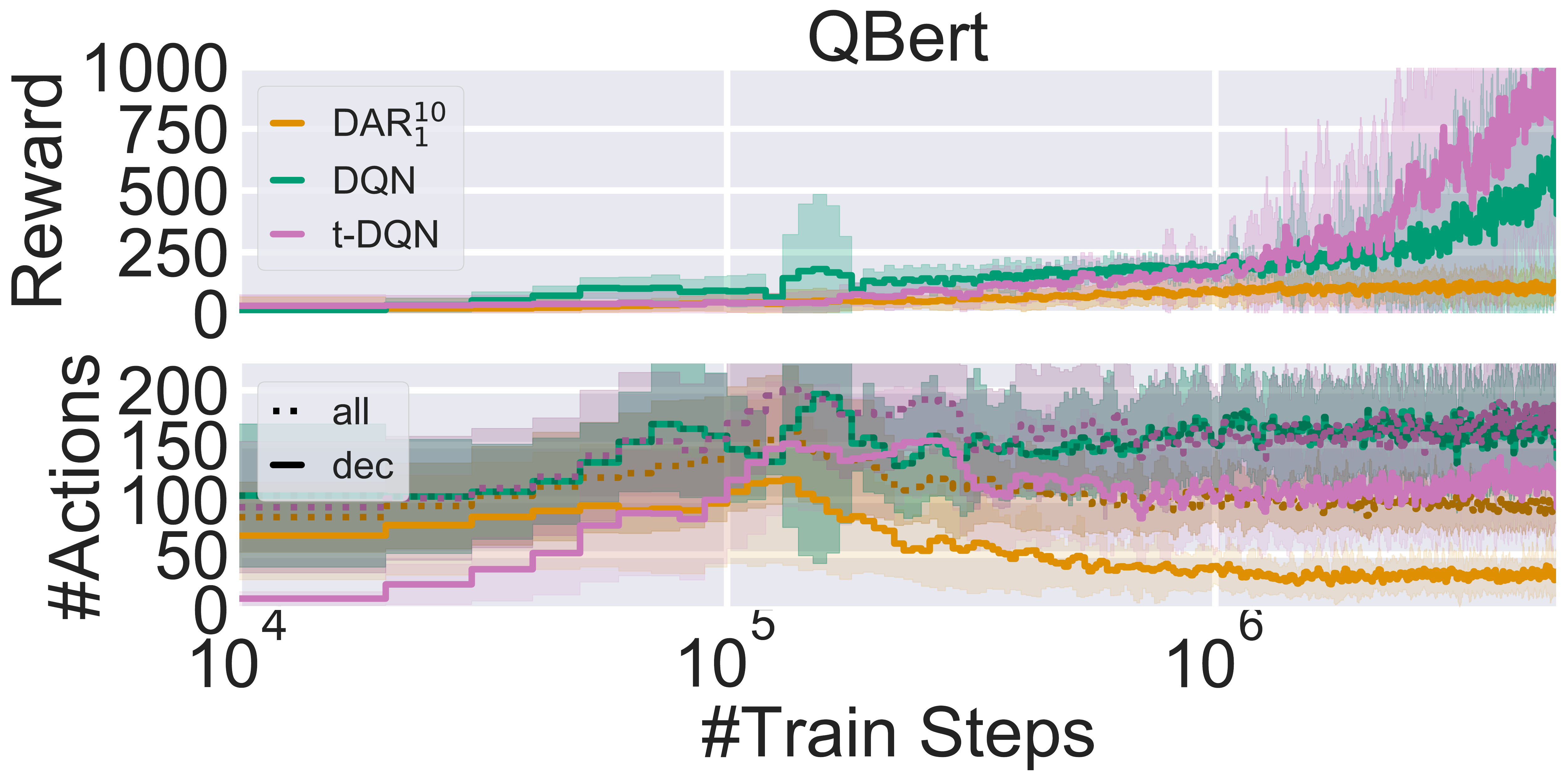}\label{fig:qbert_perf}}
    \subfloat{\centering
    \includegraphics[width=.32\textwidth]{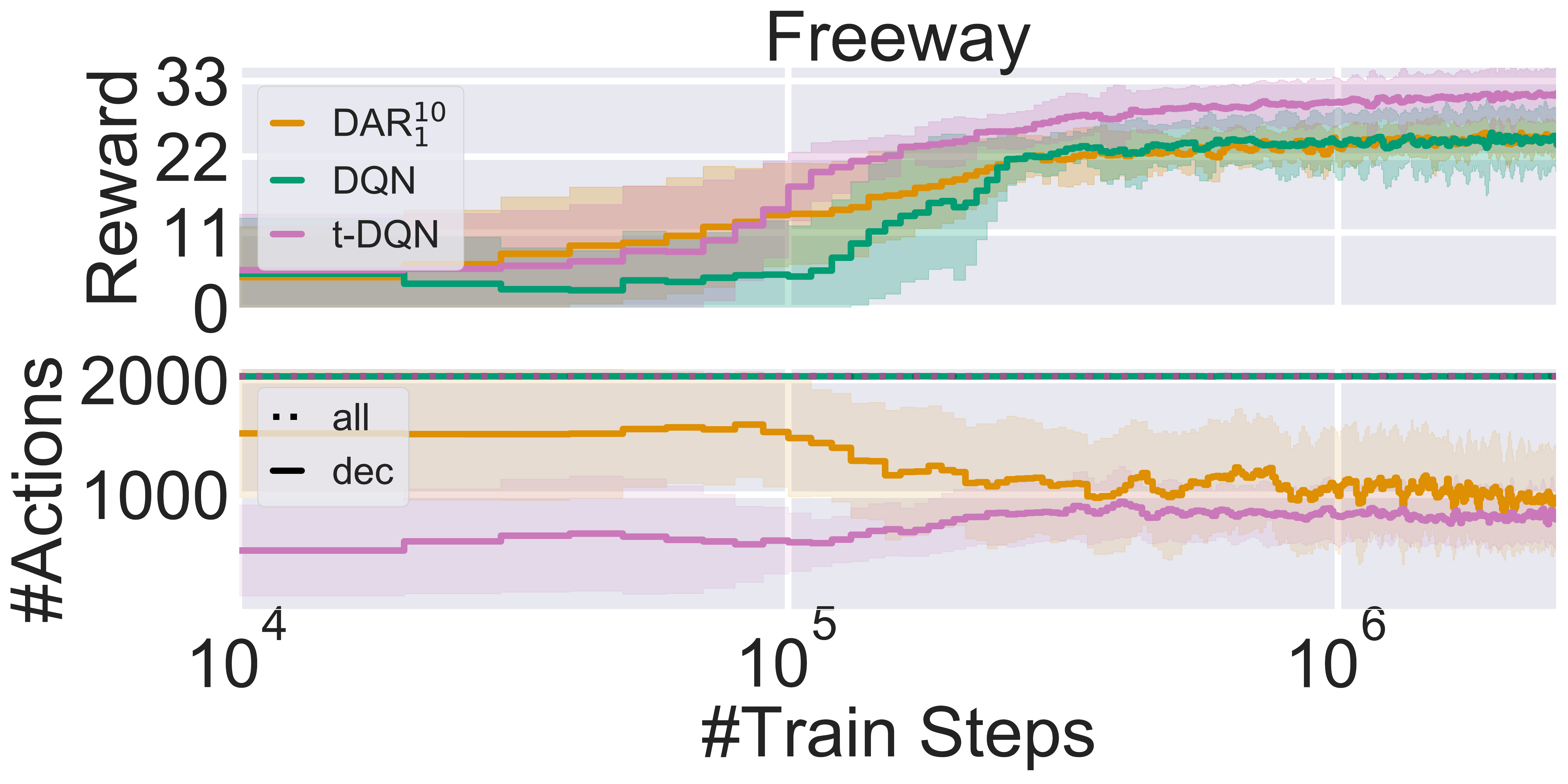}\label{fig:freeway_perf}}
    \caption{
    Evaluation performance on Atari environments. Solid lines give the mean and the shaded area the standard deviation over $15$ random seeds.
    (top) Achieved rewards. (bottom) Length of executed policy ($\cdots$) and number of decisions (---) made by the policies. To make trends more visible, we smooth over a window of width $7$.}
    \label{fig:comp_img}
\end{figure*}
\subsubsection{Atari Environments}\label{sec:experiments_atari}
\textbf{Setup} 
We trained all agents for a total of $2.5\times10^6$ training steps (i.e. only $10$ million frames) using a linearly $\epsilon$-greedy exploration schedule over the first $200\,000$ time-steps with a final fixed $\epsilon$ set to $0.01$. We evaluated all agents every $10\,000$ training steps and evaluated for $3$ episodes.
For increased learning stability we implemented all agents using double deep $\q$ networks.
For DQN we used the architecture of \citet{Mnih-nature15} which also serves as basis for our shared t-DQN and the DAR architecture.
As maximal skip-value we chose $10$.
A detailed list of hyperparameters is given in Appendix~\ref{appendix:sec:atari}.
Following~\cite{bellamare-jair13}, we used a frame-skip of $4$ to allow for a fair comparison to the base DQN. We used \textit{OpenAI Gym}'s
 We trained all agents on the games \textsc{BeamRider}, \textsc{Freeway}, \textsc{MsPacman}, \textsc{Pong} and \textsc{Qbert}.%

\textbf{Learning When to Act in Atari}
Figure~\ref{fig:comp_img} depicts the learning curves as well as the number of steps and decisions.
The training behaviour from our \temporal agents falls into one of three categories on all evaluated Atari games.

(i) Our learned t-DQN exhibits a slight improvement in learning speed, on \textsc{MsPacman} and \textsc{Pong}\footnote{Results for \textsc{Pong} and \textsc{BeamRider} are given in Appendix~\ref{appendix:sec:atari}.} before being caught up by DQN, with both methods converging to the same final reward (see Figures \ref{fig:mspc_perf} \& \ref{appendix:fig:pong_perf}).
Nevertheless, \temporal learns to make use of different degrees of fine and coarse control to achieve the same performance.
For example, a trained proactive \temporal policy requires roughly $33\%$ fewer decisions.
DAR on the other hand, learns to overly rely on the coarse control, leading to far fewer decisions but also worse final performance.

(ii) On \textsc{Qbert} the learning performance of our t-DQN lags behind that of vanilla DQN over the first $10^6$ steps.
Figure~\ref{fig:qbert_perf} (bottom) shows that in the first $\approx0.5\times10^6$ steps, \temporal first has to learn which skip values are appropriate for Qbert.
In the next $\approx0.5\times10^6$ steps, our t-DQN begins to catch up in reward, while using its learned fine and coarse control,  before starting to overtake its vanilla counterpart.
As it was not immediately clear if this trend would continue after $2.5\times10^6$ training steps, we continued the experiment for twice as many steps.
\temporal continues to outperform its vanilla counterpart, having learned to trade off different levels of coarse and fine control.
The effect of over-reliance of DAR on the coarse control is further amplified on \textsc{Qbert} resulting in far worse policies than either vanilla DQN and \temporal.

(iii) In games such as \textsc{Freeway} and \textsc{BeamRider} (Figures \ref{fig:freeway_perf} \& \ref{appendix:fig:beam_perf}), we see an immediate benefit to jointly learning \emph{when} and \emph{how} to act through \temporal.
For these games, our t-DQNs begin to learn faster and achieve a better final reward than vanilla DQNs.
An extreme example for this is \textsc{Freeway}, where the agents have to control a chicken to cross a busy multi-lane highway as often as possible within a fixed number of frames.
To this end one action has to be played predominantly, whereas the other two possible actions are only needed to sometimes avoid an oncoming car.
The vanilla DQN learns to nearly constantly play the predominant action, but does not learn proper avoidance strategies, leading to a reward of $\approx25$ (i.e successfully crossing the road $25$ times).
t-DQN on the other hand not only learns faster to repeatedly play the predominant action, but also learns proper avoidance strategies by learning to anticipate \emph{when} a new decision has to be made, resulting in an average reward close to the best possible reward of $34$.
Here, DAR profits from the use of a coarse control, learning faster than vanilla DQN.
However, similarly to vanilla DQN, DAR learns a policy that can only achieve a reward of $25$, not learning to properly avoid cars.

%--------------------------------------------------------------------------------------------
%--------------------------------------------------------------------------------------------
%--------------------------------------------------------------------------------------------
\section{Analysis of TempoRL Policies}

\begin{figure}[tbp]
    \vskip -.25cm
    \centering
    \subfloat[MountainCar]{
    \includegraphics[width=0.2\textwidth]{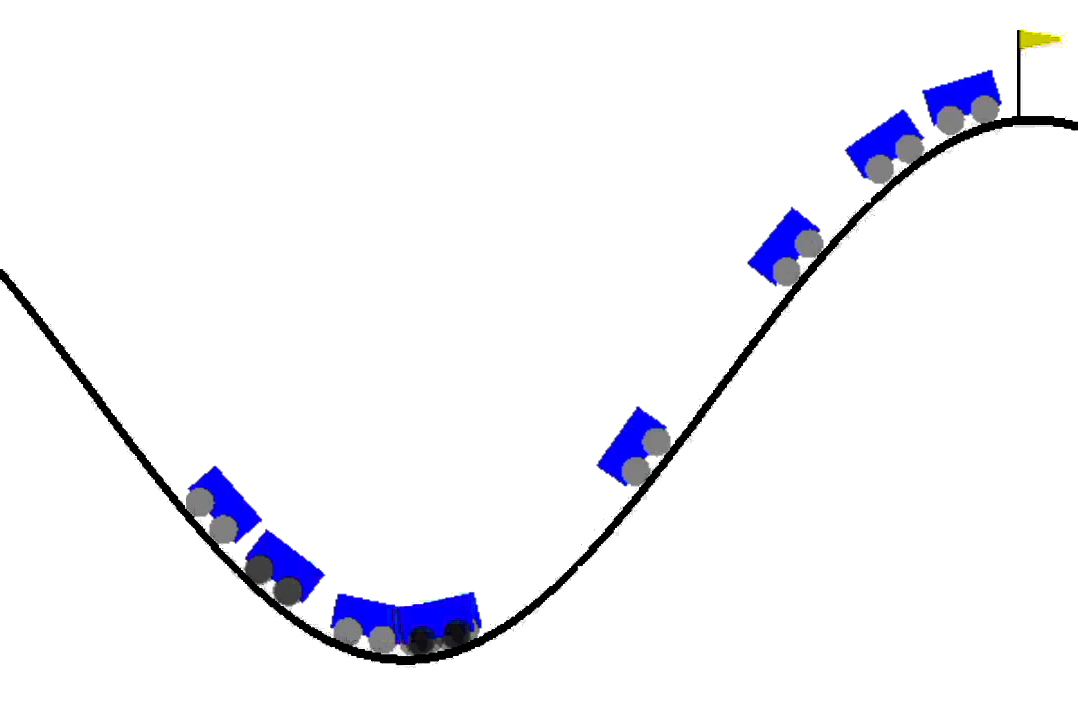}\label{fig:mountain_when}}
    \subfloat[QBert]{
    \includegraphics[width=0.2\textwidth]{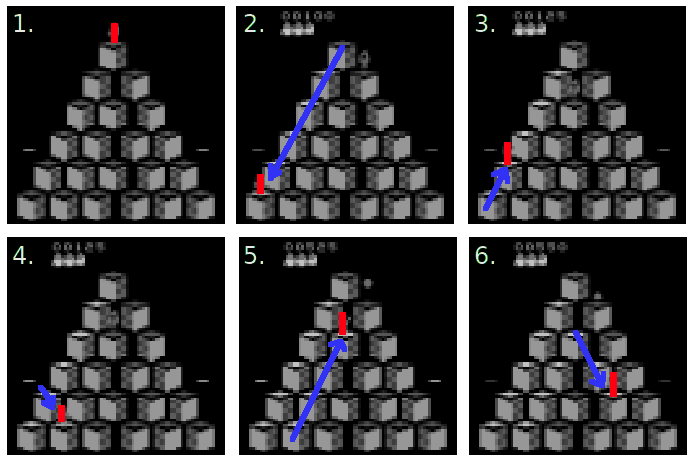}\label{fig:qbert_when}}
    \caption{
    Example States in which \temporal makes new decisions. The agents are trained with a maximal skip-length of $10$. \protect\subref{fig:mountain_when} Example states in which \temporal learned when to make new decisions in MountainCar, starting slightly to the right of the valley. \protect\subref{fig:qbert_when} Example states of Qbert. To make it easier to see where QBert is in the images we highlight him as a red square and indicate the taken trajectory with a blue arrow.}
    \label{fig:my_label}
    \vskip -.25cm
\end{figure}
To analyze \temporal policies and the decisions \emph{when} to act we selected trained agents and evaluated their policies on the environments they were trained on.
Videos for all the behaviours we describe here are part of the supplementary\footnote{Available at \href{https://www.github.com/automl/TempoRL}{\nolinkurl{github.com/automl/TempoRL}}}.
In the tabular case we plot the key-states for an agent (see Figure~\ref{fig:grids}) that can skip at most $7$ steps ahead. On the Cliff environment \temporal learns to make a decision in the starting state, once it has cleared the cliff, once before reaching the other side (since it can not skip more than $7$ states) and once to go down into the goal. Similar observations can be made for all grid-worlds. This shows that in this setting our \temporal agents are capable of skipping over unimportant states and learned \emph{when} they are required to perform novel actions.

Key-states in which \temporal decides to take new actions in the featurized MountainCar environment are shown in Figure~\ref{fig:mountain_when}.
Starting slightly to the right of the valley, the agent learns to gain momentum by making use of skips, repeating the left action\footnote{
Note, in the particular example given in Figure~\ref{fig:mountain_when} the agent first performs the left action twice, each for one time-step before it recognizes that it is gaining momentum and it can make use of large skips.}. As soon as \temporal considers the run-up to be sufficient to clear the hill on the right, it switches the action direction. From this point on \temporal sticks with this action and always selects the largest available skip-length (i.e. $10$). Still, \temporal has to make many intermediate decisions, as the agent is limited by the maximal skip-length.

Finally, we evaluated \temporal's skipping behaviour on Qbert.
An example of key states in which \temporal decides to make new decisions are given in Figure~\ref{fig:qbert_when}.
Our \temporal agent learns to use large skip-values to reach the bottom of the left column, lighting up all platforms in between. After that the agent makes use of large skips to light up the second diagonal of platforms. Having lit up a large portion of the platform,  \temporal starts to make fewer uses of skips. This behaviour is best observed in the video provided in the supplementary.
Also, note that we included all trained networks in our supplementary such that readers can load the networks to study their behaviour.

This analysis confirms that \temporal is capable of not only reacting to states but also learning to anticipate \emph{when} a switch to a different action becomes necessary.
Thus, besides the benefit of improved training speed through better guided exploration, \temporal improves the interpretability of learned policies.
%--------------------------------------------------------------------------------------------
%--------------------------------------------------------------------------------------------
%--------------------------------------------------------------------------------------------
\section{Conclusion}
We introduced skip-connections into the existing MDP formulation to propagate information about future rewards much faster by repeating the same action several times.
Based on skip-MDPs, we presented a learning mechanism that makes use of existing and well understood learning methods.
We demonstrated that our new method, \temporal is capable of learning not only how to act in a state, but also \emph{when} a new action has to be taken, without the need for prior knowledge.
We empirically evaluated our method using tabular and deep function approximation and empirically evaluated the learning behaviour in an adversarial setting.
We demonstrated that the improved learning speed not only comes from the ability of repeating actions but that the ability to learn which repetitions are helpful provided the basis of learning \emph{when} to act.
For both tabular and deep RL we demonstrated the high effectiveness of our approach and showed that even in environments requiring mostly fine-control our approach performs well.
Further, we evaluated the influence of exploration strategies, architectural choices and maximum skip-values of our method and showed it to be robust.

As pointed out by \citet{huang-allerton19}, observations might be costly.
In such cases, we could make use of \temporal to learn how to behave and when new actions need to be taken; when using the learned policies, we could use the learned skip behaviour to only observe after having executed the longest skips possible.
All in all, we believe that \temporal opens up new avenues for RL methods to be more sample efficient and to learn complex behaviours.
As future work, we plan to study distributional \mbox{\temporal} as well as how to employ different exploration policies when learning the skip policies and behaviour policies.

\section*{Acknowledgements} The authors acknowledge support by the state of Baden-W\"{u}rttemberg through bwHPC,  André Biedenkapp, Raghu Rajan and Frank Hutter by the German Research Foundation (DFG) through grant no INST 39/963-1 FUGG as well as funding by the Robert Bosch GmbH, and Marius Lindauer by the DFG through LI 2801/4-1. The authors would like to thank Will Dabney for providing valuable initial feedback as well as Fabio Ferreira and Steven Adriaensen for feedback on the first draft of the paper.%

\bibliography{bib/strings,bib/lib,bib/local,bib/proc}
\bibliographystyle{icml2021}

\appendix
\onecolumn
\renewcommand{\thetable}{\Alph{section}\arabic{table}}
\renewcommand{\thefigure}{\Alph{section}\arabic{figure}}

\section{Detailed Baseline Description}\label{appendix:sec:baselines}%
\setcounter{table}{0}%
\setcounter{figure}{0}%

\paragraph{Dynamic Action Repetition}
\citep[DAR;][]{lakshminarayanan-aaai17} is a framework for
\vskip -.075in
\begin{wrapfigure}{r}{0.23\textwidth}%
	\vskip -.7in%
	\centering%
	\scalebox{.6}{%
		\begin{tikzpicture}[shorten >=1pt,->,draw=black!50, node distance=.5cm]%
		\node (a) {$s_{0,t}$};
\node (b) [above of=a] {$s_{1,t}$};
\node (c) [above of=b] {$\vdots$};
\node (d) [above of=c] {$s_{N-1,t}$};
\node (e) [above of=d] {$s_{N,t}$};

\node (inBL) [right=.75cm of a.south] {};
\node (inUR) [right=2.25cm of d.north] {};
\node (inUUR) [right=2.25cm of e.north] {};
\path [activity] (inBL) rectangle (inUUR);
\node (inC) [rotate=90, align=center] at ($(inBL)!0.5!(inUUR)$) {Shared Feature\\Representation};

\draw[myarrow] (a) -- (inBL |- a);
\draw[myarrow] (b) -- (inBL |- b);
\draw[myarrow] (c) -- (inBL |- c);
\draw[myarrow] (d) -- (inBL |- d);
\draw[myarrow] (e) -- (inBL |- e);

\node (hiddenBL) [above right=-.3cm and 2.75cm of c] {};
\node (hiddenUR) [above right=.125cm and 1.5cm of inUUR] {};
\path [activity] (hiddenBL) rectangle (hiddenUR);
\node (hiddenC) [rotate=90, align=center] at ($(hiddenBL)!0.5!(hiddenUR)$) {Action\\Output};

\node (hiddenAUR) [below=1.55cm of hiddenUR] {};
\node (hiddenABL) [below=1.55cm of hiddenBL] {};
\path [activity] (hiddenABL) rectangle (hiddenAUR);
\node (hiddenAC) [rotate=90, align=center] at ($(hiddenABL)!0.5!(hiddenAUR)$) {Action\\Output};

% From D to Skip
\draw[myarrow] (inUR |-d) -- ($(inUR |- d)+(0.125,0)$) -- ($(inUR |- e)+(.5,.35)$) -- ($(inUR |- e)+(.7125,.35)$);
% From C to Skip
\draw[myarrow] (inUR |-c) -- ($(inUR |- c)+(0.125,0)$) -- ($(inUR |- e)+(.5,0.05)$) -- ($(inUR |- e)+(.7125,.05)$);
% From B to Skip
\draw[myarrow] (inUR |-b) -- ($(inUR |- b)+(0.125,0)$) -- ($(inUR |- d)+(.5,.2)$) -- ($(inUR |- d)+(.7125,.2)$);

% From D to Act
\draw[myarrow] ($(inUR |- d)+(0.125,0)$) -- ($(inUR |- b)+(.5,-.15)$) -- ($(inUR |- b)+(.7125,-.15)$);
% From C to Act
\draw[myarrow] ($(inUR |- c)+(0.125,0)$) -- ($(inUR |- a)+(.5,0)$) -- ($(inUR |- a)+(.7125,0)$);
% From B to Act
\draw[myarrow] ($(inUR |- b)+(0.125,0)$) -- ($(inUR |- a)+(.5,-.3)$) -- ($(inUR |- a)+(.7125,-.3)$);

\node (cc) at ($(hiddenBL |- c)+(2,.9)$) {$a_{t,r_{1}}$};
\draw[myarrow,very thick] (hiddenUR |-cc) -- (cc);

\node (bb) at ($(hiddenABL |- c)+(2,-.9)$) {$a_{t,r_{2}}$};
\draw[myarrow,very thick] (hiddenAUR |-bb) -- (bb);%
		\end{tikzpicture}%
	}%
	\caption{Schematic DAR Architecture with duplicate output heads to learn at two time-scales $r_1$ and $r_2$.}%
	\label{fig:dar_net}%
	\vskip -.1in%
\end{wrapfigure}
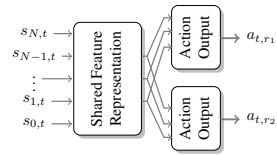%
discrete-action space deep RL algorithms.
For a discrete-action space $\mathcal{A} = \left\{a_1,\ldots,a_{|\mathcal{A}|}\right\}$ DAR duplicates this space such that an agent can choose from $2\times|\mathcal{A}|$ actions.
Further, DAR introduces two hyperparameters $r_1$ and $r_2$, each of which are associated with one half of the new action space.
These hyperparameters determine the number of time-steps an action will be played for, with both actions $a_k$ and $a_{2k}$ ($1\leq k\leq|\mathcal{A}|$) performing the same behaviour but $a_k$ is repeated for $r_1$ time-steps and $a_{2k}$ for $r_2$ time-steps.
When training an agent, there are no modifications to the training procedure, other than an agent now having to select from a larger action space.
Figure~\ref{fig:dar_net} schematically depicts a DAR DQN agents $\q$-network architecture.

This gives an agent two levels of control to decide on how long to apply an action.
A drawback of this framework is that the output heads are independent from each other and are not aware that certain action outputs have the same influence on the environment for $\min(r_1, r_2)$ time-steps.
Further, both $r_1$ and $r_2$ have to be defined beforehand, requiring good prior knowledge about the potential levels of fine and coarse control in an environment.
% Lastly, DAR requires to modify an existing agent to be usable.
% Although the modification is simple, this could lead to unexpected learning behaviour of the base agent.

\paragraph{Fine Grained Action Repetition} \citep[FiGAR;][]{sharma-iclr17} is a framework for both discrete and continuous action spaces.
Instead of learning a single policy that has to learn both which action to play and how long to follow it (as in DAR), FiGAR decouples the behaviour and repetition learning by using two separate policies $\pi_a\colon\states\to\actions$ and $\pi_r\colon\states\to\left\{1, 2, \ldots, \textrm{max repetition}\right\}$.
When training an agent, based on a state $s$, $\pi_a$ decides which action to play and simultaneously $\pi_r$ decides how long to repeat a selected action starting from $s$.
At the time of selecting their respective actions, neither $\pi_a$ nor $\pi_r$ are aware of the other policies decision.
Thus, the action and the respective repetition value are selected independently from each other.

To couple the learning of both policies \citet{sharma-iclr17} use a joint loss to update the network weights and further suggest to use weight-sharing of the input-layers of the two policy networks.
Although this aligns the policies when performing a training step, at decision time the policies remain uninformed about each others behaviour.
Counter to DAR, FiGAR allows for much more fine-grained control over the action repetition.
However, FiGAR requires more modification of a base algorithm to allow for learning of control at different time-steps.
With \temporal we propose a method that allows for the same fine-grained level of control while requiring no modifications to the base agent architecture.

% \chngd{
% \paragraph{Differences to TempoRL}
% We argue that the similarities between TempoRL and prior approaches are mostly superficial. One crucial difference is that both DAR and FiGAR require invasive modifications to the base agent.
% TempoRL on the other hand does not need to modify the learning behaviour of the base agent
% and can be readily applied to any value-based RL agent. 
% We gave an example for this in the DDPG case where we used DQN to learn the skip-policy based on the actor behaviour of an unmodified DDPG agent. 
% Further,
% prior works do not learn \textit{when} to act. DAR can only select between two levels of temporal abstraction (requiring fine tuning),
% and FiGAR is only capable of learning which repetition value works well on average for all actions (which makes it fail for Pendulum).
% Further, skip-behaviour and action behaviour are not directly aligned in prior methods, leading to sometimes competing behaviour.
% TempoRL facilitates cooperation between skip and action behaviour by conditioning the former on the latter.
% }

\clearpage
\setcounter{figure}{0}
\setcounter{table}{0}
\section{Implementation Details}\label{appendix:sec:implement}
Algorithm~\ref{alg:example} details how to train a \temporal \qlearning agent.
All elements that are new to \temporal are shown in black whereas vanilla \qlearning code is greyed out.
The functions $td\_update$ (Line 12) and $td\_update\_skip$ (Line 18) are formally stated in Equations \ref{eq:tdu} and \ref{eq:tdsu} respectively and give the temporal difference updates required during learning.
\begin{algorithm}[tb]
	\caption{\temporal \qlearning}
	\label{alg:example}
	\begin{algorithmic}[1]
		\STATE {\bfseries Input:} {\color{gray}environment $env$ with states $\mathcal{S}$ and actions $\mathcal{A}$}, {\color{black} skip-Actions $\mathcal{J}$},
		
		\hspace{1.1cm}{\color{gray}behaviour} and {\color{black} skip $\q$-functions} {\color{gray}$\q(\cdot, \cdot)$}, {\color{black} $\q(\cdot, \cdot| \cdot)$}, {\color{gray}training episodes $E$}

		\STATE Initialize ${\color{gray}\q(s, a)}, {\color{black} \q(s, j| a)}\forall {\color{gray}s\in\mathcal{S}, a\in\mathcal{A}}, {\color{black} j\in\mathcal{J}}$
		{\color{gray}
			\FOR{episode$\in\left\{1,\ldots,E\right\}$}
			\STATE $s \leftarrow env.reset()$
			\REPEAT
			\STATE $a\leftarrow\pi(s)$ \COMMENT{e.g. $\epsilon\egargmax_{a'\in\mathcal{A}}\q(s,a')$}
			{\color{black}\STATE  $j\leftarrow\pi_j(s, a)$ \COMMENT{e.g. $\epsilon\egargmax_{j'\in\mathcal{J}}\q(s,j'|a)$}
				\STATE trajectory $\leftarrow \left[s\right]$ \COMMENT{Tracks the skip trajectory}}
			{\color{black} \REPEAT
				{\color{gray}\STATE $r, s' \leftarrow env.step(a)$}
				\STATE append $s'$ to trajectory \COMMENT{Records the state transitions}
				{\color{gray}\STATE $\q(s, a)\leftarrow td\_update(\q(s, a), r, s')$\COMMENT{See Equation~\ref{eq:tdu}}
					\STATE $s \leftarrow s'$}\label{alg:line:tdu}
				\UNTIL{all skips $1,\ldots, j$ performed or episode ends}
				\STATE $\mathcal{G} \leftarrow build\_connectedness\_graph($trajectory$)$ \LONGCOMMENT{Build a local connectedness graph from the observed trajectory}
				\FORALL{connections $c\in\mathcal{G}$}
				\STATE get $s_{start}, s_{end}, j', r'$ from $c$
				\STATE $\q(s_{start}, j' | a)\leftarrow td\_update\_skip(\q(s_{start}, j' | a), r', s_{end})$\COMMENT{See Equation~\ref{eq:tdsu}}
				\ENDFOR
			}
			\UNTIL{episode finished}
			\ENDFOR
		}
	\end{algorithmic}
\end{algorithm}

\begin{equation}\label{eq:tdu}\vspace{-.5cm}
\q(s_t, a_t) = \q(s_t, a_t) +  \alpha\left(\underbrace{\left(\underbrace{r_{t} + \gamma \max\q(s_{t+1}, \cdot)}_\text{TD-Target}\right) - \q(s_t, a_t)}_\text{TD-Delta}\right)
\end{equation}
\begin{equation}\label{eq:tdsu}
\q(s_t, j_t | a_t) = \q(s_t, j_t | a_t) +  \alpha\left(\underbrace{\left(\underbrace{\sum_{k=0}^{j-1}\gamma^{k} r_{t+k} + \gamma^{j} \max\q(s_{t+\skipaction}, \cdot)}_\text{TD-Target}\right) - \q(s_t, j_t | a_t)}_\text{TD-Delta}\right)
\end{equation}

Where $\alpha$ is the learning rate and $\gamma$ the discount factor.
Note that the TD-Target in Equation~\ref{eq:tdsu} (as well as the skip $\q$-function in Equation~\ref{eq:skip}) is using the behaviour $\q$-function and not the skip $\q$-function. 
Thus, the skip $\q$-function estimates the expected future rewards, assuming that the current skip will be the only skip in the MDP.
This allows us to avoid overestimating $\q$-values through multiple skips and focuses on learning of the value of the executed skip similar to double \qlearning \cite{hasselt-neurips10}.
Further, learning of the skip-values does not interfere with learning of the behaviour $\q$-function.

The function $build\_connectedness\_graph$ (Line 15) builds takes an observed trajectory and builds connectedness graph of states that are reachable by repeatedly playing the same action (see Figure~\ref{fig:example_skip} in the main paper). Each connection contains information about start and end states, the length of the skip and the discounted reward for that skip.

\clearpage
\section{Used Compute Resources}\label{appendix:sec:compute}
\paragraph{Tabular \& Deep RL Experiments on Featurized Environments} For the tabular as well as the deep experiments on featurized environments, we evaluated all agents on a compute cluster with nodes equipped with two Intel Xeon Gold 6242 32-core CPUs, 20 MB cache and and 188GB (shared) RAM running Ubuntu 18.04 LTS 64 bit.
In all cases, the agents were allocated one CPU.
The tabular agents required at most $20$ minutes to complete training, whereas the deep agents required at most $15$ hours.

\paragraph{Deep RL Experiments on Atari Environments} These experiments were run on a compute cluster with nodes equipped with two Intel Xeon E5-2630v4 and 128GB memory running CentOS 7.
For training, the agents were allocated $10$ CPUs and required at most $48$ hours to complete training.

% \clearpage
\section{Gridworld Details}\label{appendix:sec:grids}
\begin{figure}[htbp!]
	\centering
	\subfloat[Cliff]{
		\centering
		\scalebox{1}{
			\begin{tikzpicture}
			\input{tikz/cliff-grid}
			\draw[line width=0.75mm,my_magenta] (0,0) to[out=150,in=210, looseness=0.375] (0,1.5)  to[out=150,in=120, looseness=0.325]  (4.5,1.5) to[out=30,in=-30, looseness=0.375] (4.5,0);
			\draw[fill=my_green] (0,0) circle (0.125cm);
			\draw[fill=my_green] (0,0.5) circle (0.125cm);
			\draw[fill=my_green] (4.5,0.5) circle (0.125cm);
			\draw[fill=my_green] (0,1) circle (0.125cm);
			\draw[fill=my_green] (4.5,1) circle (0.125cm);
			\draw[line width=0.75mm,my_green,dashed] (0,0) -- (0,1.5) -- (4.5,1.5) -- (4.5,0);
			\foreach \x in {0,0.5,...,4.5} {
				\draw[fill=my_green] (\x, 1.5) circle (0.125cm);
			}
			\draw[fill=my_magenta] (0,0) circle (0.0625cm);
			\draw[fill=my_magenta] (0,1.5) circle (0.0625cm);
			\draw[fill=my_magenta] (4.5,1.5) circle (0.0625cm);
			\node at (4.5,0) {\includegraphics[scale=0.125]{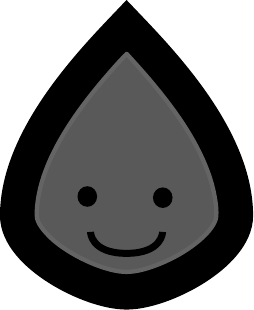}};
			\end{tikzpicture}
		}\label{fig:cliff_copy}
	}%
	\subfloat[Bridge]{
		\centering
		\scalebox{1}{
			\begin{tikzpicture}
			\input{tikz/bridge-grid}
			\draw[line width=0.75mm,my_magenta] (0,0) to[out=150,in=210, looseness=0.375] (0,1)  to[out=150,in=120, looseness=0.325]  (4.5,1) to[out=30,in=-30, looseness=0.375] (4.5,0);
			\draw[fill=my_green] (0,0) circle (0.125cm);
			\draw[fill=my_green] (0,0.5) circle (0.125cm);
			% \draw[fill=my_green] (0,1) circle (0.125cm);
			\draw[fill=my_green] (4.5,0.5) circle (0.125cm);
			% \draw[fill=my_green] (4.5,1) circle (0.125cm);
			\draw[line width=0.75mm,my_green,dashed] (0,0) -- (0,1) -- (4.5,1) -- (4.5,0);
			\foreach \x in {0,0.5,...,4.5} {
				\draw[fill=my_green] (\x, 1) circle (0.125cm);
			}
			\draw[fill=my_magenta] (0,0) circle (0.0625cm);
			\draw[fill=my_magenta] (0,1) circle (0.0625cm);
			\draw[fill=my_magenta] (4.5,1) circle (0.0625cm);
			\node at (4.5,0) {\includegraphics[scale=0.125]{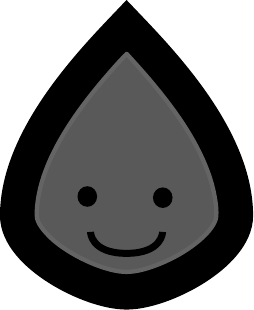}};
			\end{tikzpicture}
		}\label{fig:bridge_copy}
	}%
	\subfloat[ZigZag]{
		\centering
		\scalebox{1}{
			\begin{tikzpicture}
			\input{tikz/zigzag-grid}
			\draw[line width=0.75mm,my_magenta] (0,0) 
			to[out=150,in=210, looseness=0.375] (0,2)  
			to[out=150,in=120, looseness=0.325]  (2.5,2) 
			to[out=30,in=-30, looseness=0.375] (2.5,.5)
			to[out=-150,in=-120, looseness=0.375] (4.5,.5)
			to[out=-150,in=-210, looseness=0.375] (4.5,2.5);
			\draw[fill=my_green] (0,0) circle (0.125cm);
			\draw[fill=my_green] (0,0.5) circle (0.125cm);
			\draw[fill=my_green] (0,1) circle (0.125cm);
			\draw[fill=my_green] (0,1.5) circle (0.125cm);
			% \draw[fill=my_green] (4.5,0.5) circle (0.125cm);
			% \draw[fill=my_green] (4.5,1) circle (0.125cm);
			\draw[line width=0.75mm,my_green,dashed] (0,0) -- (0,2) -- (2.5,2) -- (2.5,.5) -- (4.5,.5) -- (4.5,2.5);
			\foreach \x in {0,0.5,...,2.5} {
				\draw[fill=my_green] (\x, 2) circle (0.125cm);
			}
			\foreach \x in {2.5,3,...,4.5} {
				\draw[fill=my_green] (\x, .5) circle (0.125cm);
			}
			\foreach \y in {1,1.5,2,2.5} {
				\draw[fill=my_green] (4.5, \y) circle (0.125cm);
			}
			\foreach \y in {1,1.5} {
				\draw[fill=my_green] (2.5, \y) circle (0.125cm);
			}
			\draw[fill=my_magenta] (0,0) circle (0.0625cm);
			\draw[fill=my_magenta] (0,2) circle (0.0625cm);
			\draw[fill=my_magenta] (2.5,2) circle (0.0625cm);
			\draw[fill=my_magenta] (2.5,0.5) circle (0.0625cm);
			\draw[fill=my_magenta] (4.5,0.5) circle (0.0625cm);
			\draw[fill=my_magenta] (4.5,2.5) circle (0.0625cm);
			\node at (4.5,2.5) {\includegraphics[scale=0.125, angle=180]{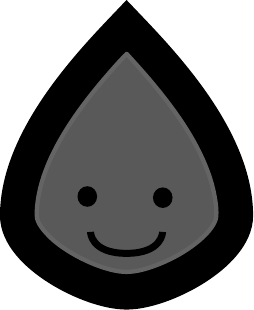}};
			\end{tikzpicture}
		}\label{fig:zigzag_copy}
	}
	\caption{Copy of Figure~\ref{fig:grids} form the main paper. $6\times10$ Grid Worlds. Agents have to reach a fixed {\color{my_blue} goal state} from a fixed {\color{my_gray} start state}. Large/small dots represent decision steps of {\color{my_green} vanilla}  and {\color{my_magenta} \temporal} \qlearning policies.}
	\label{fig:grids_copy}
\end{figure}

All considered environments (see Figure~\ref{fig:grids_copy}) are discrete, deterministic, have sparse rewards and have size $6\times10$. Falling off a cliff results in a negative reward~($-1$) and reaching a goal state results in a positive reward ($+1$). Both cliff and goal states terminate an episode.
All other states result in no reward. An agent can only execute the actions $up$, $down$, $left$, $right$ with diagonal moves not possible.
If the agent does not reach a goal/cliff in $100$ steps, an episode terminates without a reward.

For the Cliff environment, a shortest path through the environment requires $16$ steps. However, to reach the goal, decisions about unique actions are only required at $3$ time points.
The first is in the starting state and determines that action $up$ should be repeated $3$-times, the next is repeating action $right$ $10$-times and the final one is repeating action $down$ $3$-times.
Thereby, an optimal proactive policy that is capable of joint decision of action and skip length requires $\sim 80\%$ fewer decisions than an optimal reactive policy that has to make decisions in each state.
As the Bridge environment is very similar, but has a smaller cliff area below, an optimal proactive policy also requires roughly $\sim 80\%$ fewer decisions.

On the more complex ZigZag environment, an optimal policy requires $20$ steps in total to reach the goal. In this environment however, an agent has to switch direction more often. Leading to a total of $5$ required decisions. Thus in this environment an optimal proactive policy requires roughly $75\%$ fewer decisions.

\clearpage
\section{Influence of the Maximum Skip-Length}\label{appendix:skip-length}
\begin{table*}[tb!]
	\centering
	\caption{Normalized AUC for reward and average number of decision steps for varying maximal skip-lengths $J$. All agents are trained with the same $\epsilon$ schedule. $\rewards$ denotes normalized area under the reward curve and $D$ the average number of decision steps. Values are results of running $10$ random seeds. Columns $1$ and $7$ are equivalent to columns $5$ \& $6$ in Table~\ref{tab:envs}.}
	\label{tab:appendix_zigzag_ablation}
	\vskip 0.1in
	\small
	% \scalebox{0.9}{
	\subfloat[linear decaying $\epsilon$-schedule]{
		\begin{tabular}{lrrrrrrrrrrrrrrrrr}
			\toprule
			{} & \multicolumn{1}{c}{\color{my_green}$\q$} & \multicolumn{7}{r}{\color{my_magenta}$t\mhyphen\q$}\\
			\cmidrule(lr){2-2} \cmidrule(lr){3-17}
			{$J$} & \multicolumn{1}{c}{$1$} & \multicolumn{1}{c}{$2$} & \multicolumn{1}{c}{$3$} & \multicolumn{1}{c}{$4$} & \multicolumn{1}{c}{$5$} & \multicolumn{1}{c}{$6$} & \multicolumn{1}{c}{$7$} & \multicolumn{1}{c}{$8$} & \multicolumn{1}{c}{$9$} & \multicolumn{1}{c}{$10$} & \multicolumn{1}{c}{$11$} & \multicolumn{1}{c}{$12$} & \multicolumn{1}{c}{$13$} & \multicolumn{1}{c}{$14$} & \multicolumn{1}{c}{$15$} & \multicolumn{1}{c}{$16$}\\
			\midrule
			$\rewards$    & $0.57$ & $0.63$& $0.76$ & $0.87$ & $0.93$ & $0.93$ & $0.92$ & $0.91$ & $0.90$ & $0.91$ & $0.88$ & $0.87$ & $0.86$ & $0.87$ & $0.85$ & $0.84$\\
			$D$ & $83.6$ & $36.5$& $20.6$ & $13.2$ & $10.1$ & $8.3$ & $7.7$ & $7.8$ & $7.5$ & $7.4$ & $7.6$ & $7.4$ & $7.6$ & $7.6$ & $7.8$ & $7.4$\\
			\bottomrule
		\end{tabular}\label{tab:appendix_zigzag_linear}
	}
	% }
	
	% \scalebox{.9}{
	\subfloat[logarithmic decaying $\epsilon$-schedule]{
		\begin{tabular}{lrrrrrrrrrrrrrrrrr}
			\toprule
			$\rewards$  &  $0.90$ & $0.91$ & $0.93$ & $0.96$ & $0.96$ & $0.96$ & $0.96$ & $0.96$ & $0.96$ & $0.96$ & $0.96$ & $0.96$ & $0.95$ & $0.96$ & $0.95$ & $0.95$\\
			$D$ & $35.6$ & $21.7$ & $14.9$ & $11.6$ & $9.5$ & $8.6$ & $6.4$ & $6.3$ & $6.5$ & $5.9$ & $6.1$ & $6.2$ & $7.0$ & $6.8$ & $7.0$ & $6.0$\\
			\bottomrule
		\end{tabular}\label{tab:appendix_zigzag_log}
	}
	% }
	
	% \scalebox{.9}{
	\subfloat[constant $\epsilon=0.1$]{
		\begin{tabular}{lrrrrrrrrrrrrrrrrr}
			\toprule
			$\rewards$  &  $0.95$ & $0.98$ & $0.98$ & $0.99$ & $0.99$ & $0.99$ & $0.99$ & $0.99$ & $0.99$ & $0.99$ & $0.99$ & $0.99$ & $0.99$ & $0.99$ & $0.99$ & $0.98$\\
			$D$ & $27.6$ & $15.8$ & $12.0$ & $9.1$ & $8.2$ & $7.8$ & $6.8$ & $6.9$ & $6.7$ & $7.1$ & $6.6$ & $7.2$ & $6.2$ & $6.5$ & $7.0$ & $6.9$\\
			\bottomrule
		\end{tabular}\label{tab:appendix_zigzag_const}
	}
	\vskip -0.1in
\end{table*}
The maximum skip length $J$ is a crucial hyperparameter of \temporal.
A too large value might lead to many irrelevant choices which the agent has to learn to ignore; whereas a too small value might not reduce the complexity of the environment sufficiently enough, leading to barely an improvement over the vanilla counterpart.
To evaluate the influence of the hyperparameter on our method we trained various \temporal agents with varying maximal skip-lengths, starting from $2$ up to $16$.
Larger skips than $10$ will never be beneficial for the agent as the agent is guaranteed to run into a wall for some steps.
Depending on where in the environment the agent is located, smaller skip-values might allow it to quickly traverse through the environment.

Table~\ref{tab:appendix_zigzag_ablation} shows the influence of $J$ on the ZigZag environment (see Figure~\ref{fig:zigzag}).
In this environment, the largest skip value that is possible without running into a wall is $6$. 
Thus, small skip values up to $5$ quickly improve the performance over the vanilla counterpart, not only in terms of anytime performance but also in terms of required decisions.
In the case of a suboptimal exploration policy, in the form of linearly decaying $\epsilon$-greedy schedule (see Table~\ref{tab:appendix_zigzag_linear}), larger skip-values quickly lead to a decrease in anytime performance, as the agent has to learn to never choose many non-improving skip actions.

For a more suiting exploration policy, too large skip-actions do not as quickly degrade the anytime performance of our \temporal agents.
In the case of a logarithmically decaying $\epsilon$ schedule (Table~\ref{tab:appendix_zigzag_log}), we can see that skip sizes larger or equal than $12$ start to negatively influence the anytime performance, whereas with a constant $\epsilon$ schedule only a skip-size of $16$, nearly $3$ times as large as the largest sensible choice, has a negative effect.

Similar observations can be made for deep \temporal on both Pedulum, MountainCar and LunarLander, see Tables \ref{tab:auc_deep_featurized_ddpg}~-~\ref{tab:auc_deep_featurized} in the main paper.
We can see that choosing larger maximal skip-values is beneficial, up to a point, at which many irrelevant, and potentially useless choices are in the action space.
For these, \temporal first has to learn on which part of the skip-action-space to focus before really learning when new decisions need to be taken.

It is worth noting that, in the tabular case, all evaluated skip-sizes $J$ result in better anytime-performance and a lower number of required decision points compared to vanilla \qlearning, for all considered exploration strategies.
In future work, we will study how to allow \temporal to select large skip-actions without needing to learn to distinguish between many irrelevant choices.
One possible way of doing this could be by putting the skip-size on a log scale.
For example using $\log_2$ could result in only $10$ actions where a \temporal agent could skip up to $1024$ steps ahead but would still be able to exert fine control with the smaller actions.

\clearpage
\section{DDPG Implementation Details and Additional Results}\label{appendix:sec:ddpg}%
\setcounter{table}{0}%
\setcounter{figure}{0}%
\begin{figure}[t!]
	\centering
	\captionsetup[subfigure]{labelformat=empty}
	\begin{tabular}{cc}
		\subfloat{\includegraphics[width=0.49\textwidth]{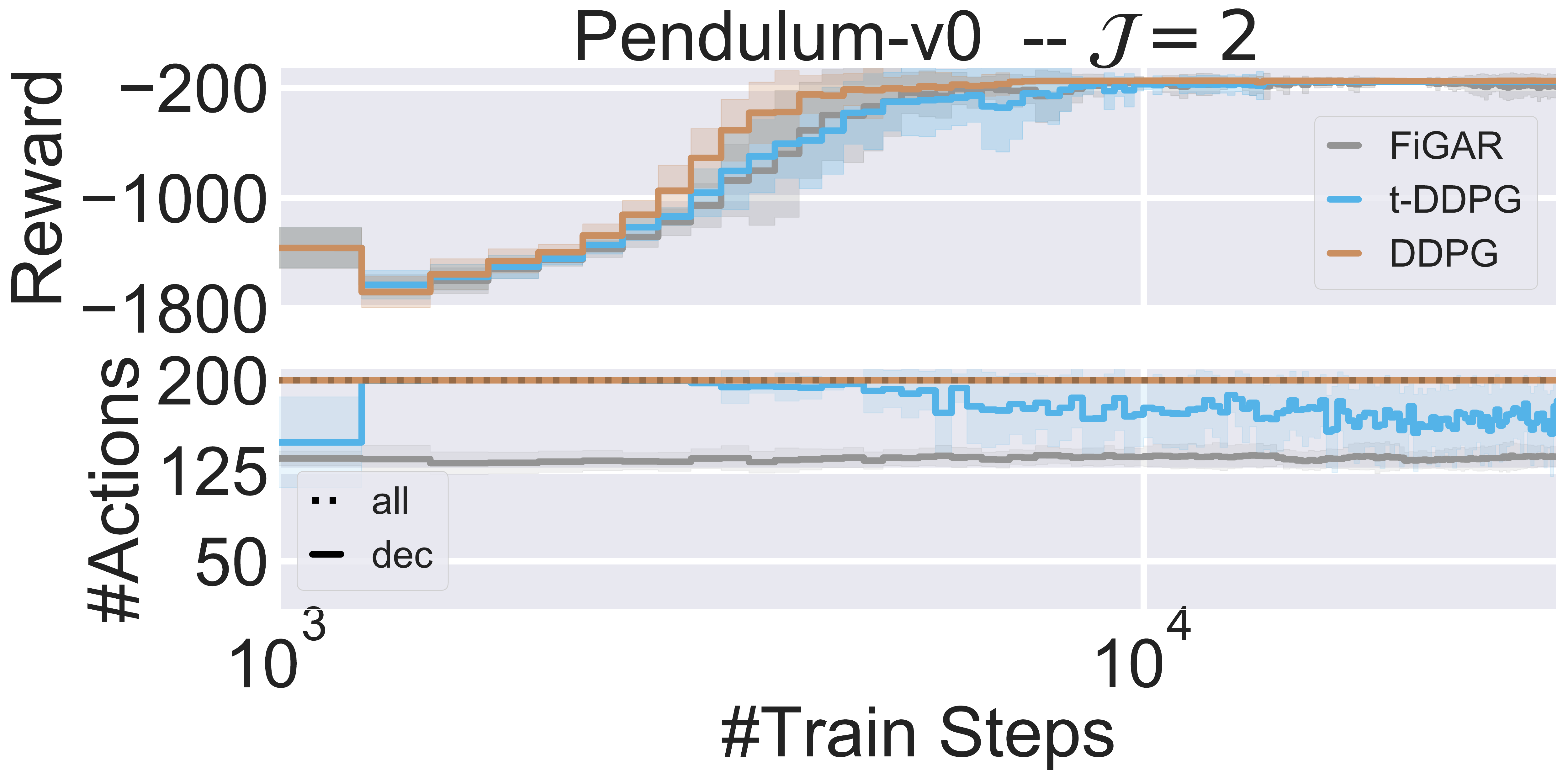}} &
		\subfloat{\includegraphics[width=0.49\textwidth]{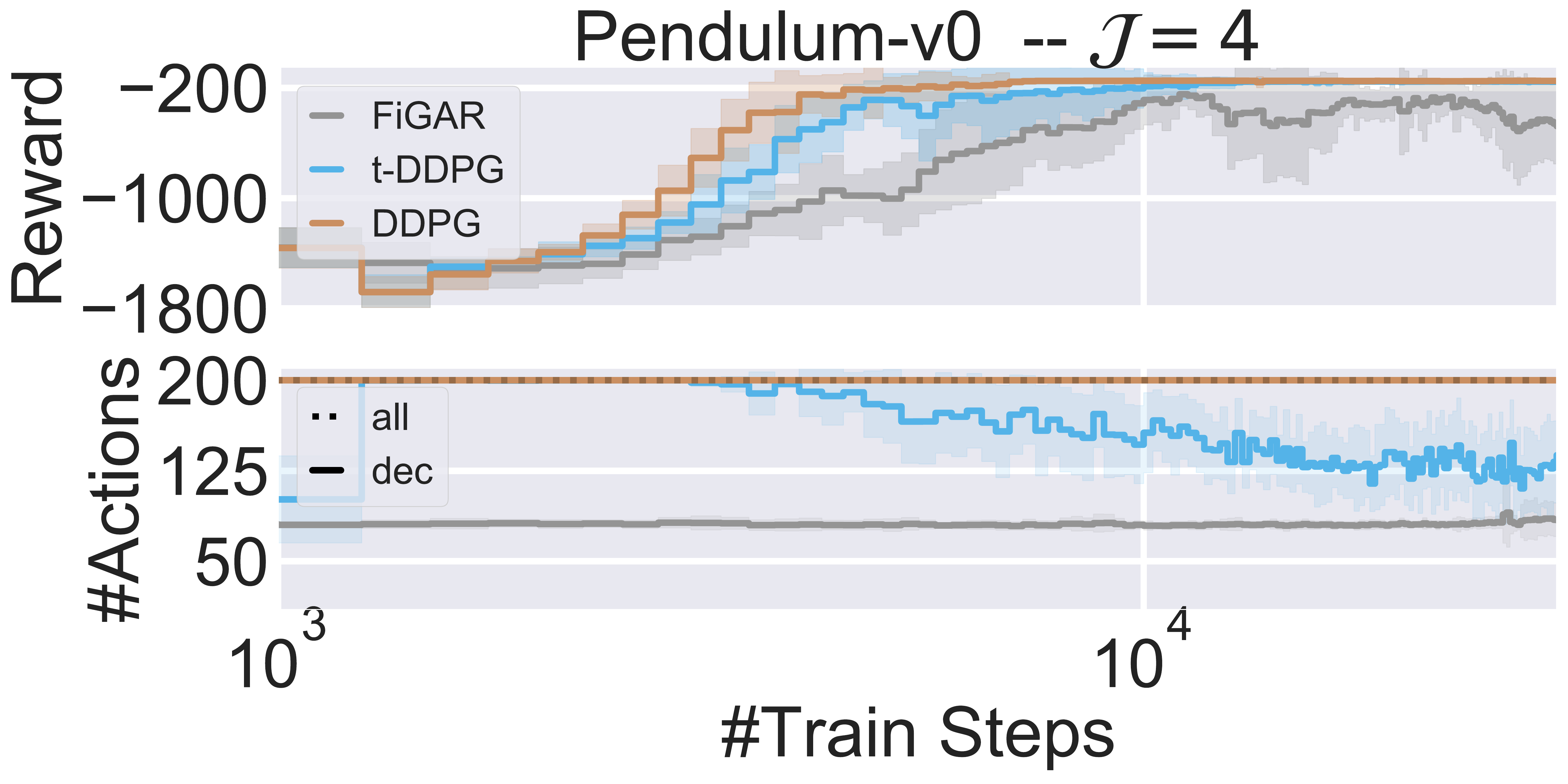}}\\
		\subfloat{\includegraphics[width=0.49\textwidth]{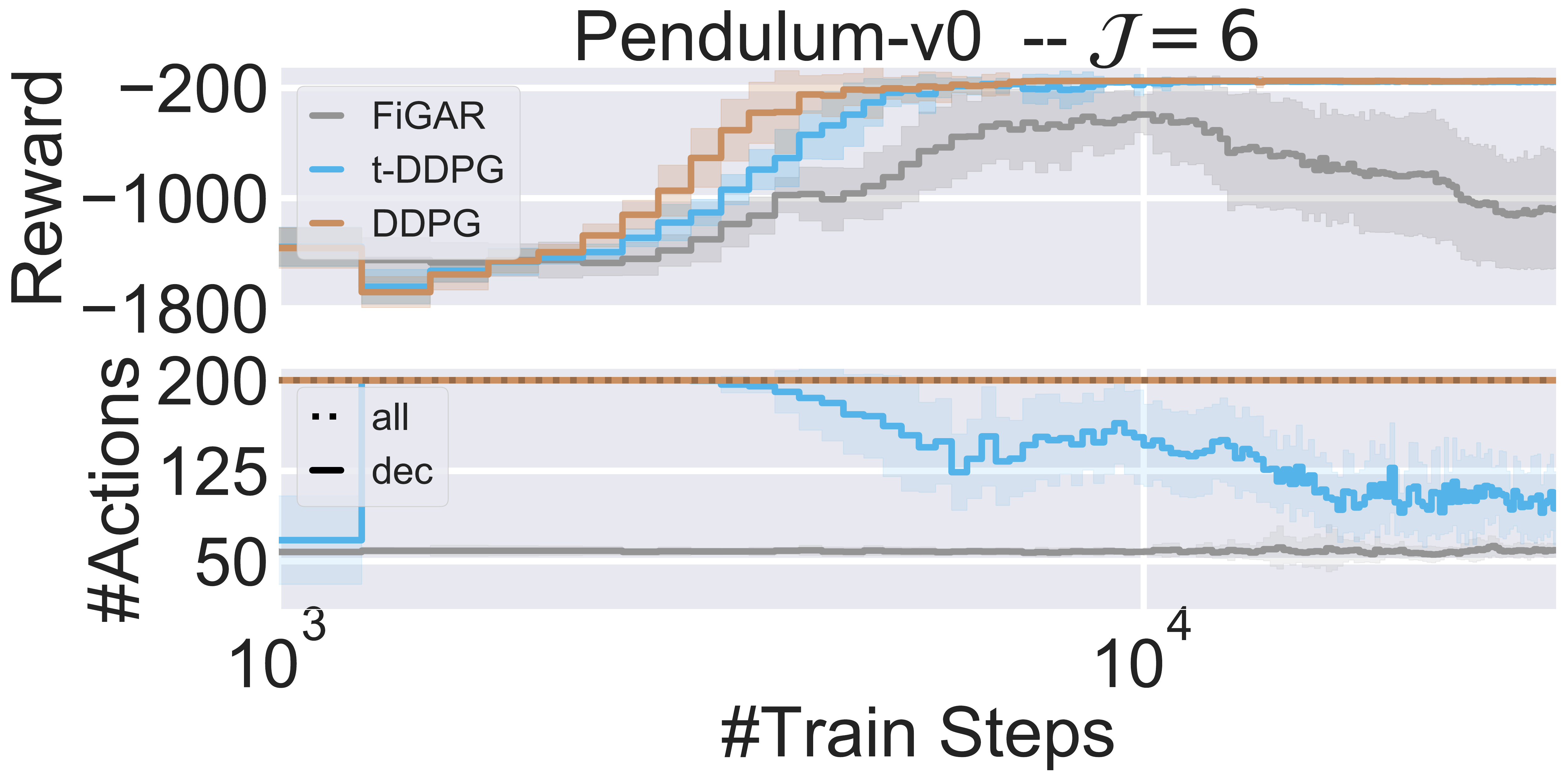}} &
		\subfloat{\includegraphics[width=0.49\textwidth]{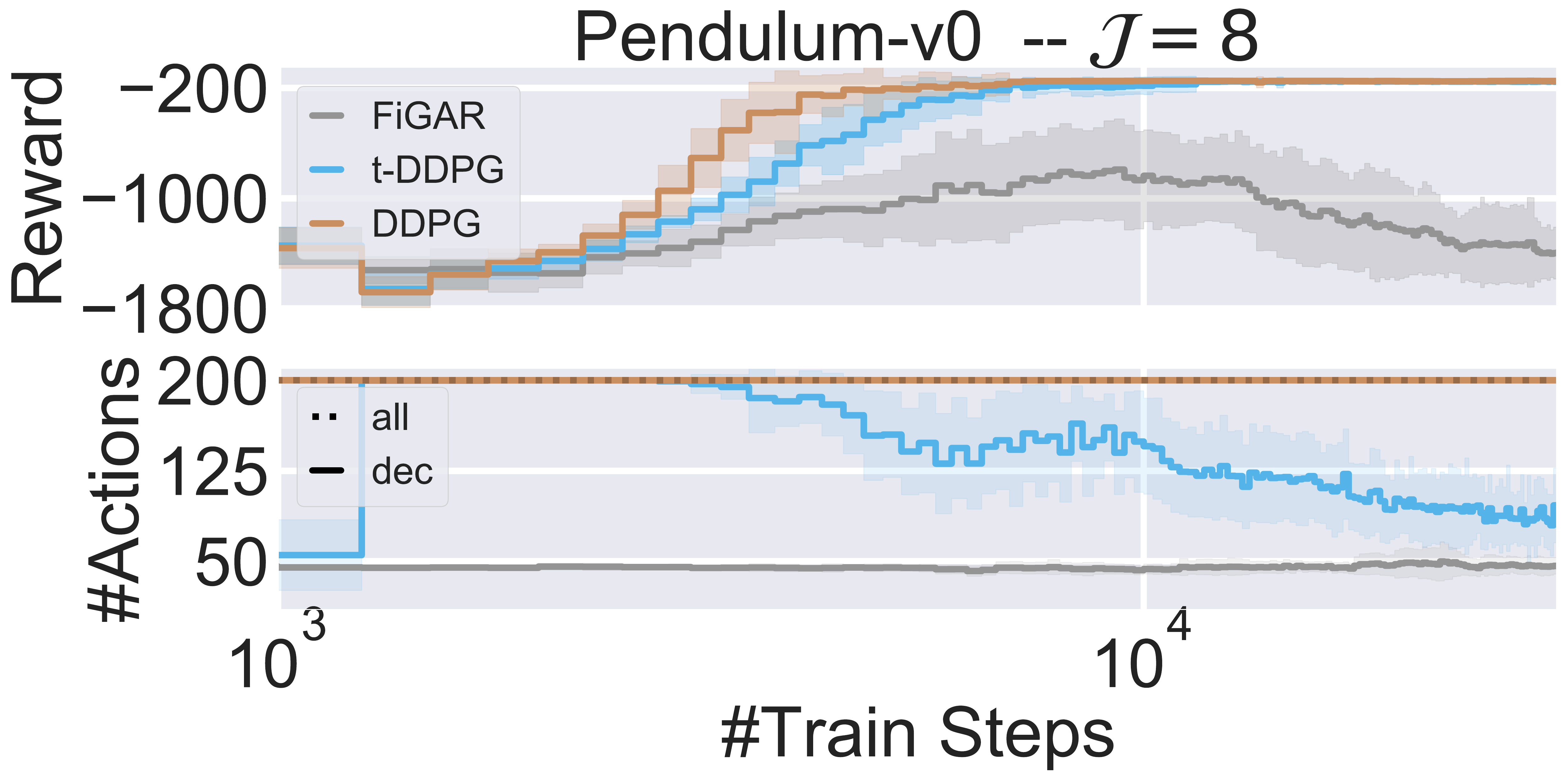}}\\
		\subfloat{\includegraphics[width=0.49\textwidth]{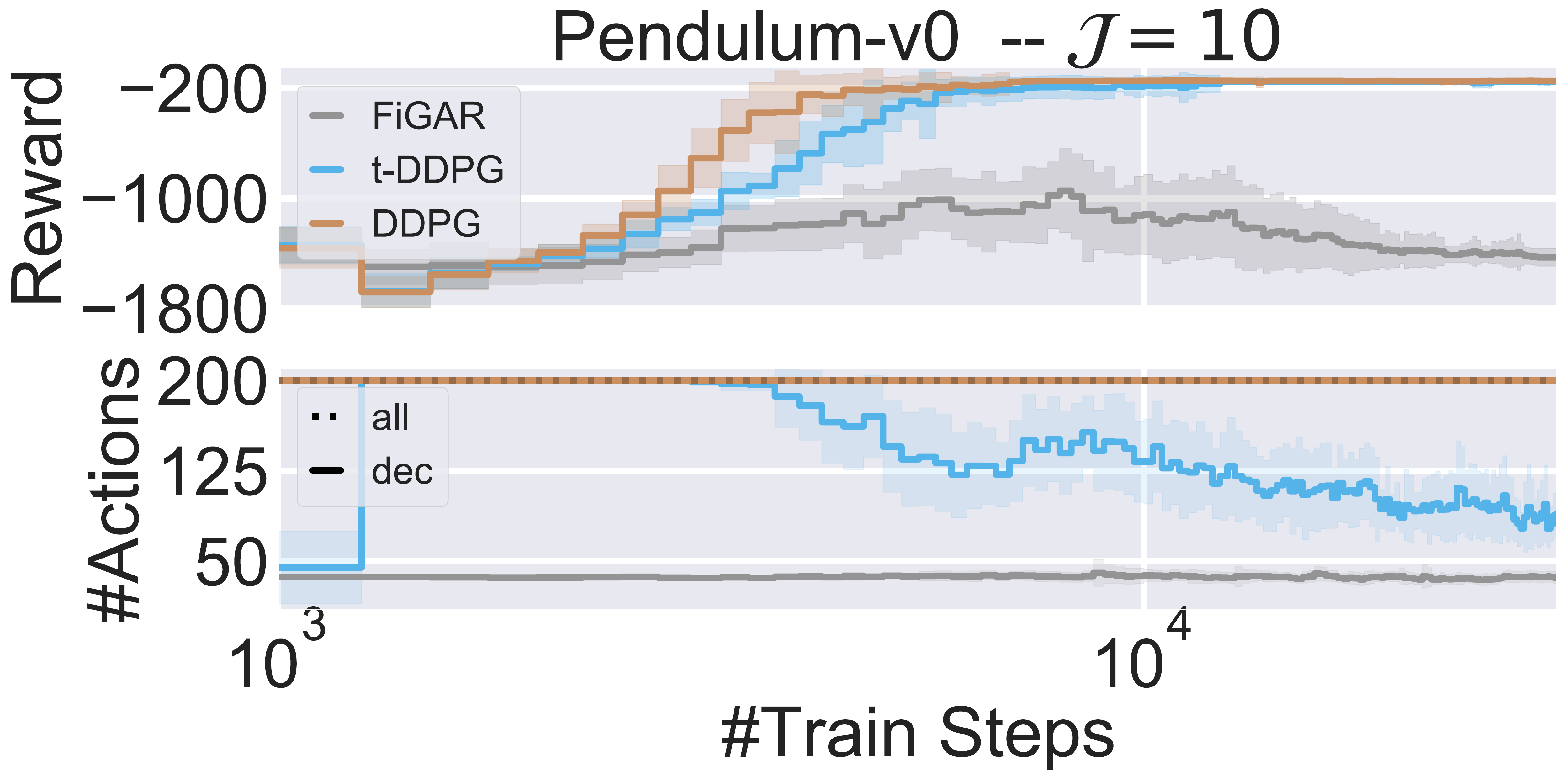}} &
		\subfloat{\includegraphics[width=0.49\textwidth]{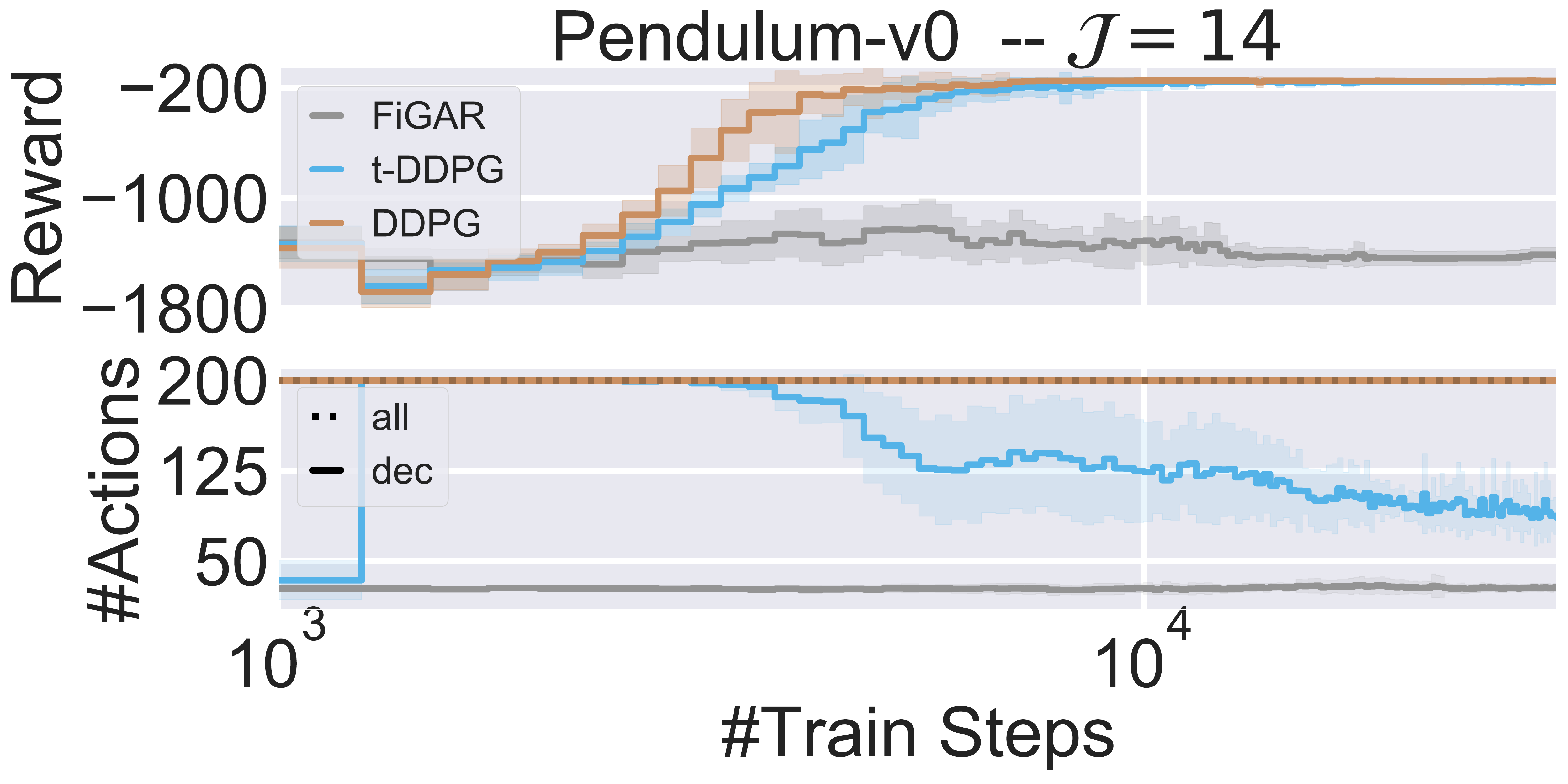}}\\
		\subfloat{\includegraphics[width=0.49\textwidth]{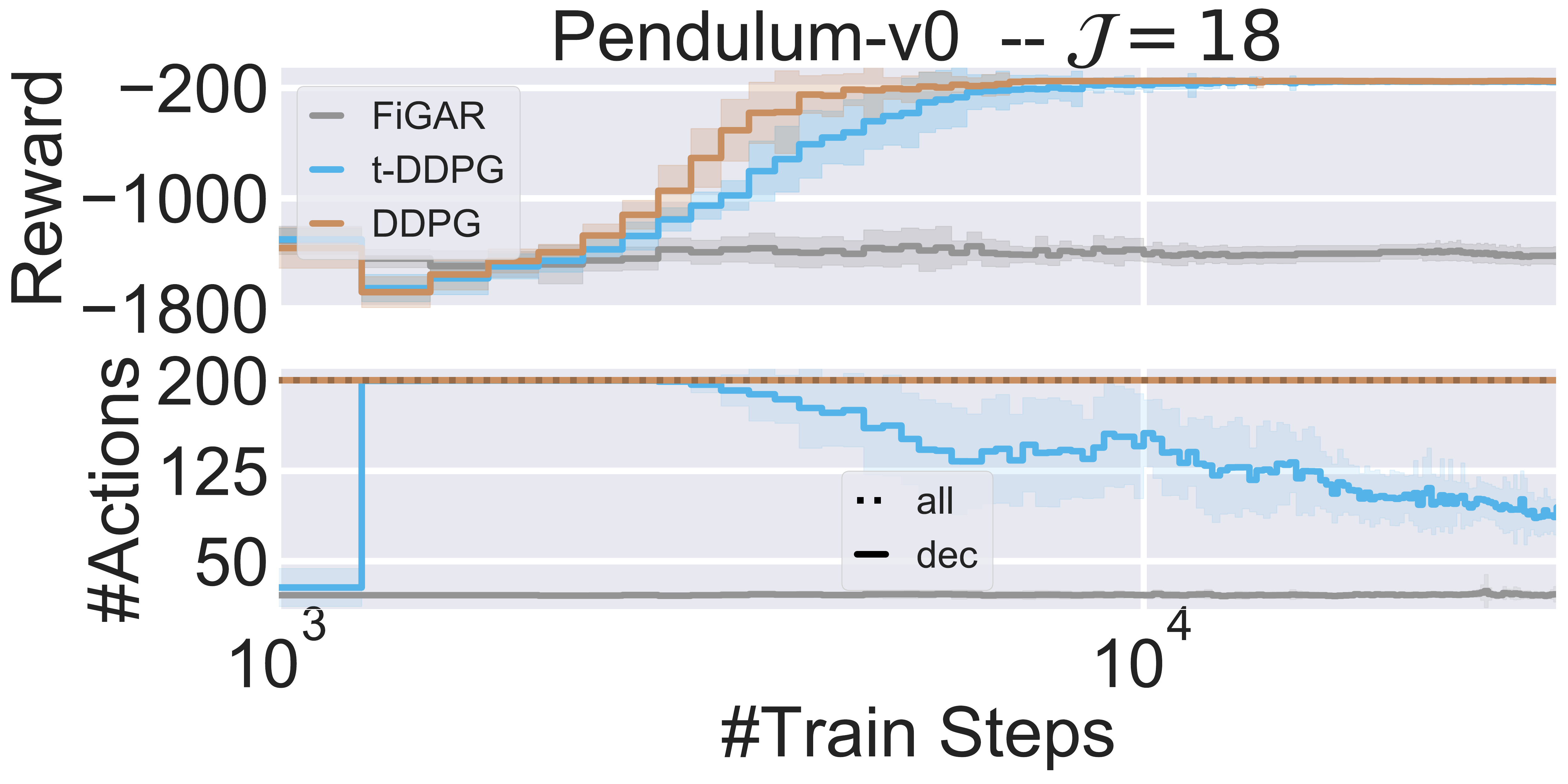}} &
		\subfloat{\includegraphics[width=0.49\textwidth]{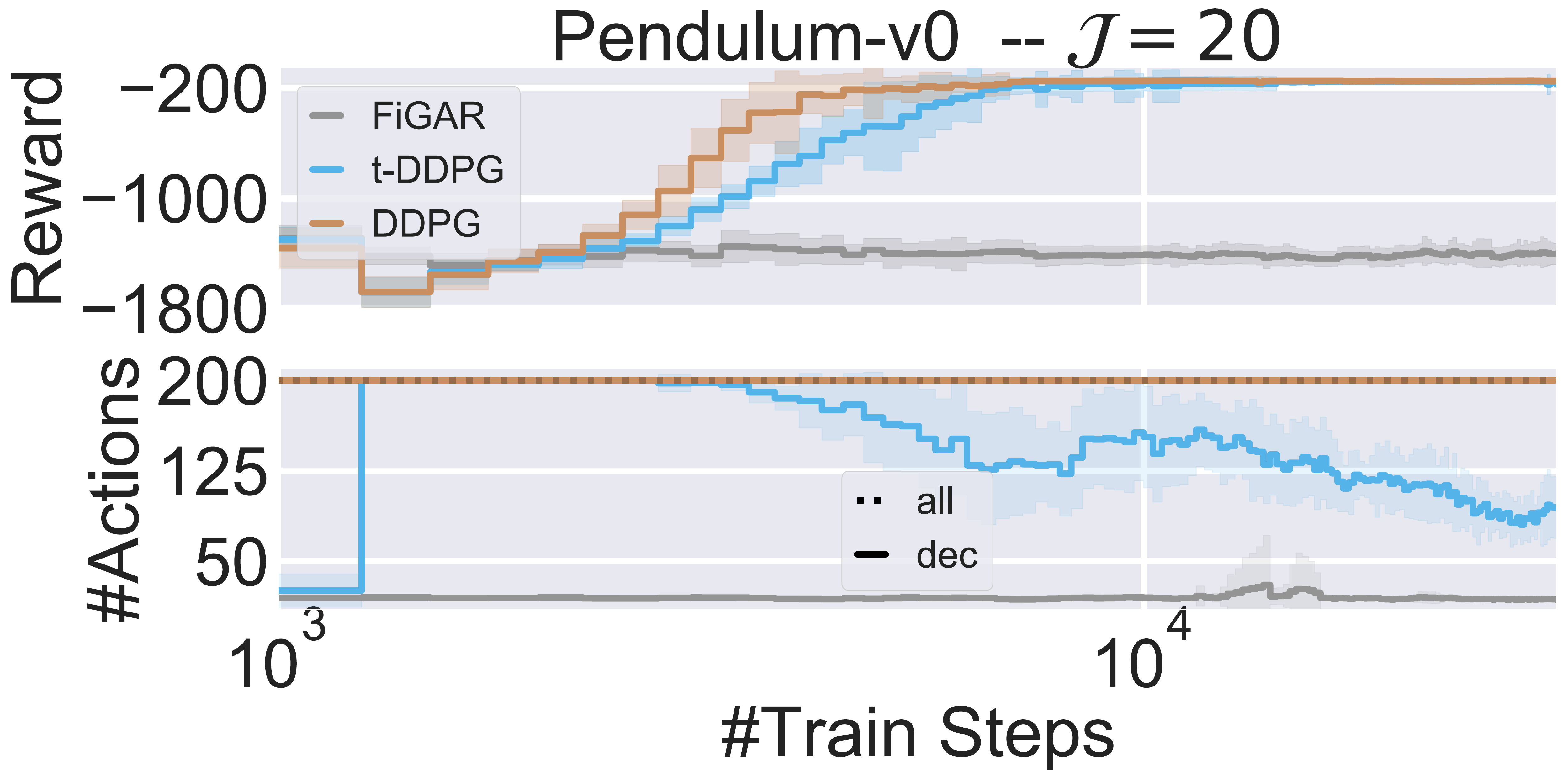}}\\
	\end{tabular}
	\caption{Learning curves of different {\color{my_ddpg_orange}DDPG} agents on Pendulum-v0. $J$ indicates the maximal skip-length used when training {\color{my_tddpg_blue}t-DDPG} and {\color{my_figar_gray}FiGAR}. Solid lines give the mean and the shaded area the standard deviation over $15$ seeds. Top-row images show the reward achieved and bottom-row images the required steps and decisions per evaluation rollout.}
	\label{appendix:fig:ddpg_all}
\end{figure}
As base implementation for DDPG, we used publicly available code\footnote{\url{https://github.com/sfujim/TD3}} and used the default hyperparameters, except we replaced the number of maximal training steps and initial  random steps as described in the main paper.
When implementing FiGAR, we followed the description by \citet{sharma-iclr17}.
Thus, the repetition policy uses a constant epsilon-greedy exploration.
Likewise, we use a constant epsilon-greedy exploration to learn our t-DDPG.

For our t-DDPG implementation we could use the same algorithm as described in Algorithm~\ref{alg:example}.
Only the greyed out parts of normal \qlearning have to replaced by DDPG training specific elements.
For example, for DDPG, the exploration policy for the actor is given by adding exploration noise rather than following an epsilon-greedy policy.
Further, we again can make use of the base agents $\q$-function as shown in Equation~\ref{eq:tdsu}.

Figure~\ref{appendix:fig:ddpg_all} depicts the learning curve for all DDPG agents with increasing maximal skip-value.
As described in the main paper, both FiGAR and t-DDPG slightly lag behind vanilla DDPG when only allowing for skips of length $2$.
However, with increasing max-skip value FiGAR quickly begins to struggle and in the end even converges to worse policies, always preferring large skip-values.
Our t-DDPG using \temporal performs much more stable and is hardly affected by increasing the maximal skip length.
Further, t-DDPG over time learns \emph{when} it is necessary to switch to new actions, roughly halving the required decisions.

% \clearpage
\section{Featurized Environments Description}\label{appendix:sec:feats}
\textbf{MountainCar} is a challenging exploration task and requires an agent to control an under powered car to drive up a steep hill on one side \cite{moore-phd90}.
To reach the goal, an agent has to build up momentum.
The agent always receives a reward of $-1$ until it has crossed the goal position and a reward of $0$ afterwards.
The observation consists of the car position and velocity and the agent can either accelerate to the left or right or do nothing.
To build up momentum an agent potentially has to repeat the same action multiple times.
Thus, we evaluate both t-DQN and DAR on the grid $\left\{2, 4, 6, 8, 10\right\}$ for the maximal (while keeping the minimal skip value fixed to $1$) skip-value over $50$ random seeds (see Tables \ref{tab:mcauc_archs} \& \ref{tab:mcauc}).

\textbf{LunarLander} The task for an agent is to land a space-ship on a lunar surface. To this end, the agent can choose to fire the main engine, steer left or right or do nothing.
Firing of the engines incurs a small cost of $-0.3$, whereas crashing or successfully landing results in a large cost or reward of $-100$ and $100$ respectively.
We expect that an environment with such a dense reward, where actions directly influence the achieved reward does not benefit from leveraging skips.

\clearpage
\section{Atari}\label{appendix:sec:atari}%
\setcounter{table}{0}%
\setcounter{figure}{0}%
\begin{figure*}[tb!]
	\centering
	\subfloat[Pong]{\centering
		\includegraphics[width=.49\textwidth]{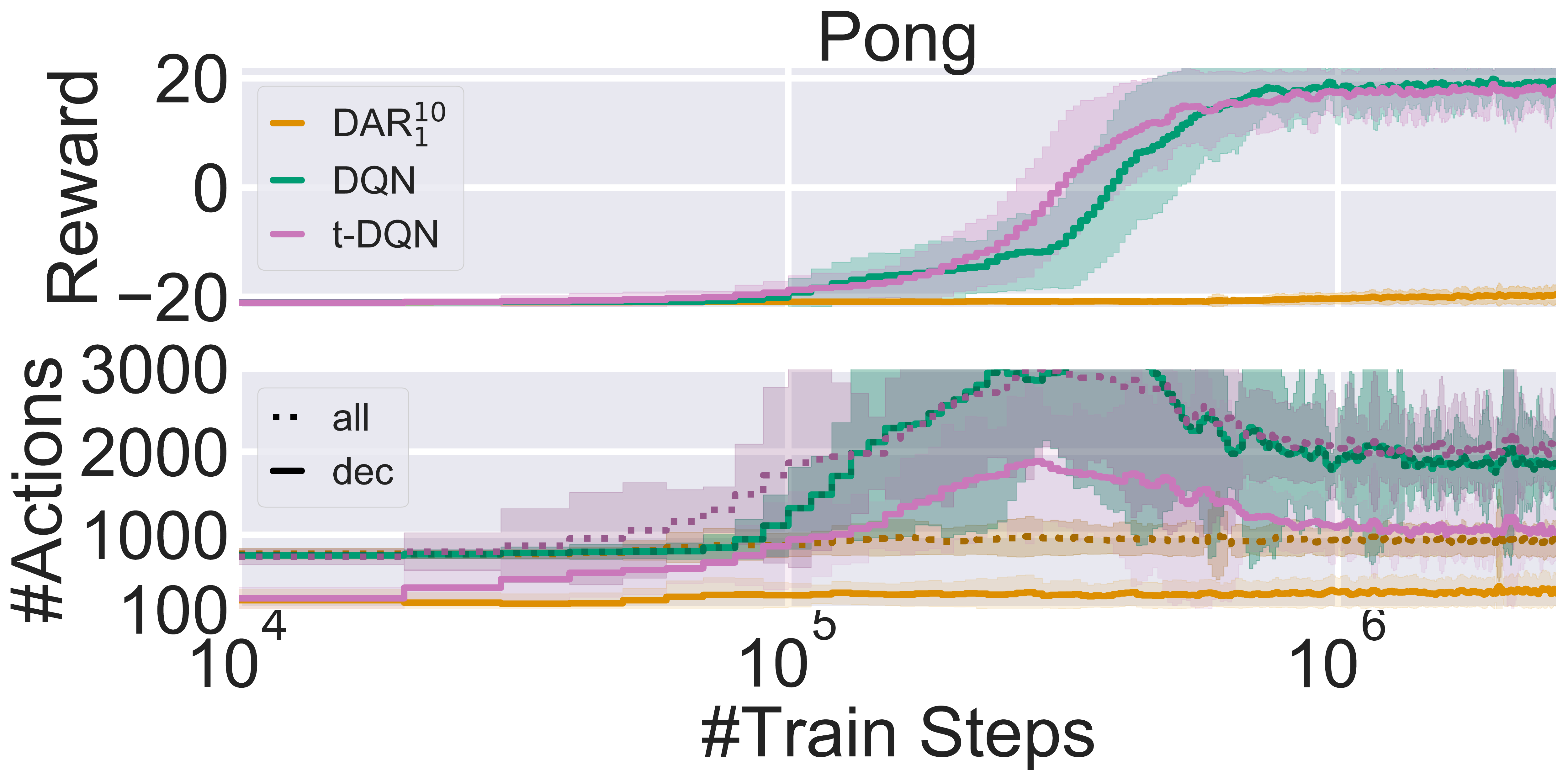}\label{appendix:fig:pong_perf}}
	\subfloat[BeamRider]{\centering
		\includegraphics[width=.49\textwidth]{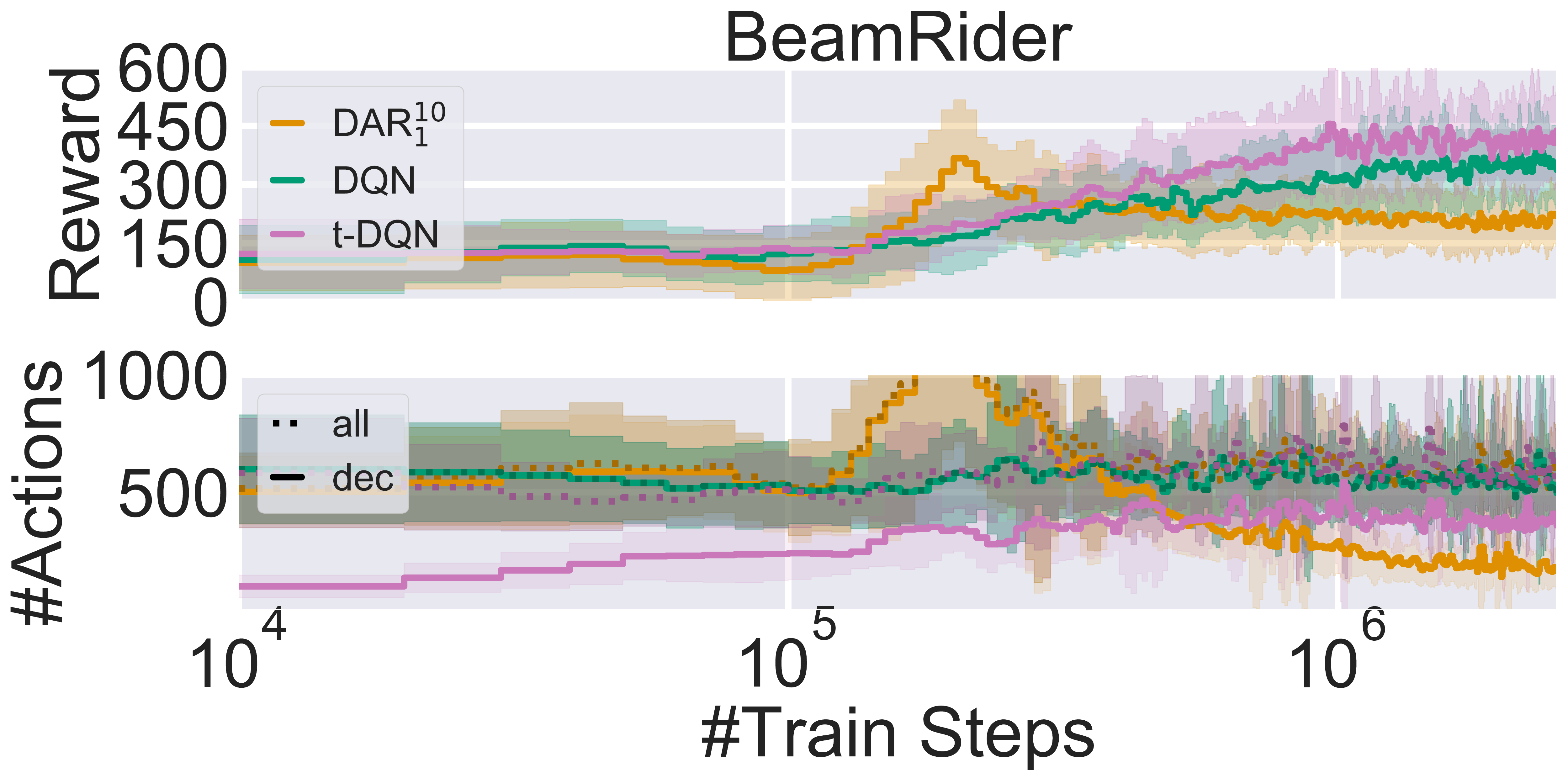}\label{appendix:fig:beam_perf}}
	\caption{
		Evaluation performance on Atari environments. Solid lines give the mean and the shaded area the standard deviation over $15$ random seeds.
		(top) Achieved rewards. (bottom) Length of executed policy ($\cdots$) and number of decisions (---) made by the policies.}
	\label{appendix:fig:comp_img}
\end{figure*}
\begin{wraptable}{r}{0.35\textwidth}%
	\vskip -1.25cm
	\centering%
	\caption{Hyperparameters used for the Atari Experiments}%
	\label{tab:my_label}%
	\begin{tabular}{l|c}%
		\toprule%
		Hyperparameter & Value \\
		\midrule
		Batch Size & $32$\\
		$\gamma$ & $0.99$\\
		Gradient Clip & $40.0$\\
		Target update frequency & $500$\\
		Learning starts & $10\,000$\\
		Initial $\epsilon$ & $1.0$\\
		Final $\epsilon$ & $0.01$\\
		$\epsilon$ time-steps & $200\,000$\\
		Train frequency & $4$\\
		Loss Function & Huber Loss\\
		Optimizer & Adam\\
		Learning rate & $10^-4$\\
		$\beta_1$ & 0.9\\
		$\beta_2$ & 0.999\\
		Replay-Buffer Size & $5\times10^4$\\
		Skip Replay-Buffer Size & $5\times10^4$\\
		$J$ & $10$\\
		\bottomrule
	\end{tabular}
	\vskip -1cm
\end{wraptable}
\paragraph{Architectures}
\emph{DQN:}
As architecture for DQN we used that of \citet{Mnih-nature15} and used this as basis for our shared architecture. This architecture has three layers of convolutions to handle the $84\times84$ input images. The first convolution layer has $84$ input channels, $32$ output channels, a kernel size of $8$ and a stride of $4$.
The second has $32$ input channels, $64$ output channels, a kernel size of $4$ and a stride of $2$.
The second has $64$ input channels, $64$ output channels, a kernel size of $3$ and a stride of $1$.
This is followed by two hidden layers with $512$ units each.

\emph{\temporal}:
The shared architecture used by our \temporal agent uses the same architecture as just described but has an additional output stream for the skip-outputs.
The skip output stream combines a hidden layer with $10$ units together with the output of the last convolutional layer.
It then processes these features again in two fully connected hidden layers with $512$ units each.

\emph{DAR:}
Similarly, the DAR agent builds on the DQN architecture of \citet{Mnih-nature15}.
However, the final output layer is duplicated and the duplicate outputs act at a different time-resolution.
To give DAR the same coarse control as would be possible with our \temporal agent we fix the fine and coarse control levels to $1$ and $10$ respectively.

\paragraph{Additional Results on \textsc{Pong}:}
Our learned t-DQN exhibits a slight improvement in learning speed, \textsc{Pong} before being caught up by DQN (similar to the results on MsPacman in the main paper, see Figure \ref{fig:mspc_perf}), with both methods converging to the same final reward.
Nevertheless, \temporal learns to make use of different degrees of fine and coarse control to achieve the same performance, requiring roughly $1\,000$ fewer decisions.

The DAR agent really struggles to learn a meaningful policy on this game, never learning to properly avoid getting scored on or scoring itself.
A likely reason for the poor performance is the choice of hyperparameters.
Potentially choosing smaller skip-value for the coarse control could allow to learn better behaviour with DAR.

\paragraph{Additional Results on \textsc{BeamRider}:} Figure~\ref{appendix:fig:beam_perf} shows an immediate benefit to jointly learning \emph{when} and \emph{how} to act through \temporal.
Our t-DQN begins to learn faster and achieve a better final reward than vanilla DQN.

Interestingly, the DAR agent, starting out with choosing to mostly apply fine control starts to learn much faster than vanilla DQN and our \temporal agent, nearly reaching the final performance of vanilla DQN already $\approx 900\,000$ time-steps earlier.
However, the performance starts to drop when DAR starts to increase usage of the coarse control.
Once the DAR agents have learned this over-reliance on the coarse control, they do not recover, resulting in the worst final performance.

\end{document}